\documentclass{article}
\usepackage{iclr2027_conference,times}

\usepackage{amsmath,amssymb}
\newcommand{\Tprof}{\mathit{Table}_{\mathrm{prof}}}   %
\newcommand{\Tscore}{\mathit{Table}_{\mathrm{score}}} %
\usepackage{booktabs}
\usepackage{float}   %
\usepackage{graphicx}
\usepackage{hyperref}
\usepackage{url}
\newif\ifibmloop
\ibmlooptrue

\title{Hybrid Quantum Selection of Learning Rates\\
 in Language-Model Training:\\
 Termination Criteria and an\\
 Interaction-Guided Schedule Oracle}

\author{Wei-Che Hung \\
Independent Researcher \\
\texttt{wolfmib@gmail.com}}

\iclrfinalcopy

\begin{document}

\maketitle
\lhead{Preprint}

\begin{abstract}
Learning-rate choices strongly affect language-model fine-tuning, and prior work shows that layer-wise learning
rates can improve training; however, the best layer-group learning-rate profile is generally unknown. This paper
studies quantum minimum finding as a layer-group learning-rate selector for a future coherent quantum training
architecture. The search combines D\"urr--H\o yer minimum finding with BBHT search and compares two classes of stopping
policies: fixed-budget and confidence-based termination. The stopping policies are first compared on a lookup oracle
built from measured profile scores at fixed training checkpoints, which reproduces the required search behavior. The
paper then introduces an Interaction-Guided Schedule Oracle, which needs no score table and works in two steps
inside one simulated circuit. First, it scores all joint schedules at once in superposition: classical one-step
loss measurements give the main effect of each layer group and the pairwise and three-way interaction effects
between groups, and the circuit adds up the effects that apply to each schedule. Second, it compares every score
with that of a baseline schedule and marks the schedules that beat it, so that amplitude amplification returns an
improving candidate; after each round of Grover iterations, the measured candidate is confirmed classically to
score better than the baseline. The baseline in this paper is uniform-rate training; a schedule proposed by any
classical adaptive method could take its place, with the quantum search looking for joint refinements beyond it, a
direction left for future research. In closed-loop fine-tuning of a 1.5-billion-parameter language model, each
selected schedule drives the next 200 training steps, and benchmark accuracies lie within the range observed across
five uniform-rate seeds. The results give a working example of how quantum superposition can be brought into
adaptive learning-rate selection during training; the limitations that remain on the quantum side are discussed in
the paper.
\end{abstract}

\section{Introduction}
Layer-wise learning rates can improve language-model training. Discriminative fine-tuning assigns
different rates across layers \citep{howard2018ulmfit}, layer-wise adaptive rate scaling does the same
for deep networks \citep{you2017lars}, and recent work derives layer-dependent rates from model
statistics or training dynamics \citep{he2026layerwise,pang2026balancing}; separate rates can also
improve the two factors of a low-rank adapter \citep{hayou2024loraplus}. These methods determine rates
individually or by a prescribed rule, rather than jointly selecting one profile from all admissible
layer-group combinations.
The number of joint profiles grows quickly with the number of groups and candidate rates: six groups with
eight admissible rates already define $8^6=262{,}144$ possible profiles.
This combinatorial growth motivates the quantum-search formulation developed here.

This paper studies quantum minimum finding for joint learning-rate selection across $G$ layer groups,
where each group chooses from $K$ admissible rates and one complete profile specifies the rates applied
to the model. The quantum search has one task: identify a low-loss learning-rate profile from these
combinations, after which the selected rates are used for the next block of classical training.
In the first part of the paper, this process is simulated at fixed training checkpoints: candidate profiles
are evaluated by short $h=25$-step replays whose held-out losses fill a score table and provide the search
scores. The checkpoints
are spaced every $200$ training steps in the experimental design; in a practical system, when to invoke
the selector could instead be determined by the training schedule or other monitored signals.

Quantum search enters only at the profile-selection step. Grover's search can identify marked candidates
in an unstructured space of $N$ possibilities using $O(\sqrt N)$ oracle queries \citep{grover1996}, and
\citet{durr1996} extend this idea to minimum finding by repeatedly lowering a threshold.
In this paper the search space contains $N=K^G$ learning-rate profiles, each represented by one basis-state
index of a quantum search register that holds all of them in superposition. For example, $N=8$ candidate
profiles require three search qubits,
represented by $|000\rangle,\ldots,|111\rangle$.
The quantum search therefore acts on profile indices and uses the profile score only to determine which
candidates improve on the current threshold.

In the score-table experiments, each search round prepares the register, marks the profiles below the current loss
threshold, amplifies them by Grover diffusion and measures; because the measurement is probabilistic, a round
may return no improving candidate even when one exists. The rounds are simulated without noise, and a small
hardware experiment verifies the marking behavior.
The marking is carried out by a lookup oracle built from measured profile scores. Run on candidate spaces of
$N=81$, $128$ and $256$, these simulations provide the data from which the performance of the termination
criteria is determined.

Quantum minimum finding proceeds through both successful and unsuccessful search rounds. A successful
round finds a better candidate and lowers the current threshold, whereas an unsuccessful round leaves
the threshold unchanged. Because the true minimum is unknown, repeated failures do not by themselves
reveal when no better profile remains.
The paper therefore compares termination criteria that decide when the search should stop: each rests on its
own assumption and stops the search once the observed rounds meet its condition. For each criterion, the work
spent finding better profiles is counted, and so is the extra effort spent after the last improvement until
the search stops.

The second part of the paper replaces the lookup oracle. The Interaction-Guided Schedule Oracle
(Section~\ref{sec:igso}) borrows a notion familiar in machine learning, main effects and interaction effects
\citep{tsang2018nid}, and brings it into superposition. Classical one-step loss measurements give the main
effect of each layer group and the interaction effects between groups; the oracle then works in two steps
inside one simulated circuit. First, the circuit computes a score for every schedule, one complete set of
group rates, in superposition. Second, it marks the schedules whose scores exceed the current threshold,
initially the score of the uniform-rate baseline. Grover diffusion then amplifies the marked schedules before
measurement. The measured candidate's score is checked classically: an improvement replaces the current best
schedule and raises the threshold; otherwise, the threshold stays unchanged. Search rounds continue, with the
rounds and Grover iterations counted, until a termination criterion stops the search. The best schedule found
drives the next block of training (Section~\ref{sec:res-closedloop}).

Machine learning already adapts learning rates during training without exhaustive search: population-based
training discovers a schedule of hyperparameter settings rather than one fixed set for the whole course of
training \citep{jaderberg2017pbt}, and hypergradient descent updates the learning rate during optimization from
the gradient of the update rule with respect to that rate \citep{baydin2018hypergradient}.
The oracle offers a further refinement step: starting from a baseline schedule, it searches for joint
changes across layer groups that achieve a higher interaction score. This paper demonstrates that process
with a uniform-rate baseline. More generally, a schedule proposed by a classical adaptive method could serve
as the baseline: new one-step measurements around that schedule would supply the main and interaction
effects, and its score would set the initial threshold. Much as a support vector machine can use different
kernels within the same learning framework \citep{cortes1995svn}, the oracle could use effects measured
around different baseline schedules while retaining the same quantum scoring and marking structure.
Extending this process to classical adaptive baselines is a direction for future research. The contributions
are a comparison of termination criteria for quantum minimum finding on measured score tables, and a
demonstration, with uniform-rate training as the baseline, in which the Interaction-Guided Schedule Oracle
searches for improved combinations of group rates and the selected rates are applied to successive blocks of
training.

\section{Methodology}

\subsection{From classical training to quantum selection}
\label{sec:object}

For scale, the original Transformer was trained for $100{,}000$ steps at about $0.4$~s per step in
its base configuration and for $300{,}000$ steps at $1.0$~s per step in its large configuration
\citep{vaswani2017attention}. LoRA fine-tuning keeps the same step-based structure while reducing
the number of trainable parameters \citep{hu2022lora}. For $n_{\mathrm{tr}}$ optimizer steps with
mean time $t_{\mathrm{tr}}$, the classical training time is
\begin{equation}
  T_C=n_{\mathrm{tr}}\,t_{\mathrm{tr}} .
  \label{eq:tc}
\end{equation}

Nonuniform learning rates have been used in several forms: discriminative fine-tuning assigns
layer-specific rates \citep{howard2018ulmfit}; layer-wise adaptive scaling treats layers separately
\citep{you2017lars}; model-derived rates use layer statistics or training dynamics
\citep{he2026layerwise,pang2026balancing}; and LoRA-specific rates assign different rates to the two
low-rank factors \citep{hayou2024loraplus}. Motivated by these results, this paper groups the
trainable layers into $G$ groups and searches only the rate $\eta_g$ assigned to each group $g$:
\begin{equation}
  w_g\leftarrow w_g-\eta_g\,d_g,\qquad g=1,\dots,G,
  \label{eq:update}
\end{equation}
where $w_g$ denotes the trainable parameters of group $g$, and $d_g$ is the update direction supplied
by back-propagation and the optimizer.

From Eq.~\eqref{eq:update}, each group chooses one learning rate. Let $\mathcal E$ contain $K$
admissible rates. A learning-rate profile $z$ selects one member of $\mathcal E$ for each of the $G$
groups, so the complete profile space contains $N=K^G$ combinations:
\begin{equation}
  \mathcal E=\{\eta^{(1)},\dots,\eta^{(K)}\},\qquad
  z=(\eta_1,\dots,\eta_G)\in\mathcal E^G,\qquad
  N=|\mathcal E^G|=K^G .
  \label{eq:profile}
\end{equation}
Each of these $N$ profiles is one candidate learning-rate setting for the model.

The quantum search register encodes the learning-rate profiles of Eq.~\eqref{eq:profile}. Let $\Tprof$
denote the table of profiles that a search covers, one profile $z$ per row and $N$ rows in total. A register
of $q_{\mathrm{srch}}$ qubits has $2^{q_{\mathrm{srch}}}$ computational basis states \citep{grover1996};
therefore a $\Tprof$ of $N$ rows requires $q_{\mathrm{srch}}=\lceil\log_2 N\rceil$ search qubits. Each
profile $z$ is represented by its binary row index as the basis state $|z\rangle$ of the register, and the
search begins from the uniform superposition over the $N$ encoded profiles:
\begin{equation}
  q_{\mathrm{srch}}=\lceil\log_2 N\rceil,\qquad
  z\;\longleftrightarrow\;|z\rangle,\qquad
  |\psi_0\rangle=\frac{1}{\sqrt N}\sum_{z}|z\rangle .
  \label{eq:encode}
\end{equation}
For the complete profile space $\mathcal E^G$, $\Tprof$ contains $K^G$ profiles. When only a subset is
searched, $N$ is the number of rows included in $\Tprof$, and Eq.~\eqref{eq:encode} fixes the width of the
search register, so the number of search qubits grows with the logarithm of $N$
(Figure~\ref{fig:object} shows the mapping at $N=8$).

To find a useful profile $z$ during training, the search over $\Tprof$ follows the quantum minimum
finding of \citet{durr1996}, which applies Grover's search \citep{grover1996} repeatedly while
lowering a threshold. Each row carries one score $f(z)$, the quantity to be minimized. Let $y$ be the
current threshold profile and $f(y)$ its score, $M$ the number of profiles in the $N$-row $\Tprof$ with
$f(z)<f(y)$, and $j$ the number of Grover iterations applied in one search round. For example, at the
scale of the held-out losses in the simulations of Section~\ref{sec:results}, if the current threshold
has $f(y)=2.70$ and exactly three profiles in $\Tprof$ have lower scores, then $M=3$; the search does not
know this value in advance. A round prepares $|\psi_0\rangle$ of Eq.~\eqref{eq:encode}, applies $j$
Grover iterations, measures once, and tests whether the measured profile lies below the current
threshold. A successful round returns some profile $z$ with $f(z)<f(y)$, which becomes the new threshold
profile. For fixed $N$, $M$, and $j$, the probability that a round returns a profile below the threshold is
\begin{equation}
  p_j(M)=\sin^2\!\big((2j+1)\theta\big),\qquad \theta=\arcsin\sqrt{M/N}.
  \label{eq:grover}
\end{equation}

\begin{figure}[!htb]
  \centering
  \includegraphics{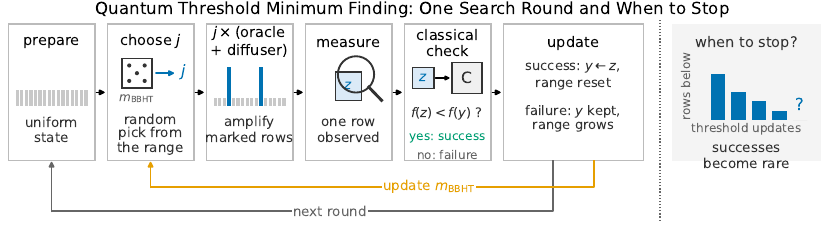}
  \caption{Quantum threshold minimum finding: one search round and when to stop. Each round prepares the
  search register, chooses an iteration count $j$, applies $j$ oracle--diffuser iterations, measures one row,
  and checks whether its score improves the current threshold $f(y)$; a success moves the threshold and resets
  the range $m_{\mathrm{BBHT}}$ from which $j$ is drawn, a failure keeps it and enlarges that range. At the
  illustrative size $N=16$ with $M=2$ rows below the threshold, a round with $j=1$ measures a marked row with
  probability $78.1\%$ by Eq.~\eqref{eq:grover}.}
  \label{fig:round}
\end{figure}

Because $M$ is unknown, the Grover iteration count that Eq.~\eqref{eq:grover} would favor is unknown
too, and too many iterations can rotate the state past the marked subspace. The exponential search of
\citet{boyer1998bbht} therefore does not fix a single Grover iteration count $j$ in advance. Instead, it
starts with a small range of possible iteration counts and widens that range after unsuccessful rounds,
so the search can reach deeper without assuming $M$ beforehand. Figure~\ref{fig:round} shows one round of this
search, and a run is a sequence of such rounds, with failures that widen the range and successes that
lower the threshold. Let $m_{\mathrm{BBHT}}$ denote the upper
scale of this range; in each round $j$ is drawn uniformly from $\{0,\dots,\lceil m_{\mathrm{BBHT}}\rceil-1\}$.
After a failed round the range expands, and after a successful threshold update it resets:
\begin{equation}
  \begin{array}{l@{\quad}l}
    \multicolumn{2}{c}{j\sim\mathcal U\{0,\dots,\lceil m_{\mathrm{BBHT}}\rceil-1\},}\\[3pt]
    \text{failure:} & m_{\mathrm{BBHT}}\leftarrow\min(\lambda m_{\mathrm{BBHT}},\sqrt N),\qquad \lambda=\tfrac65,\\
    \text{success:} & m_{\mathrm{BBHT}}\leftarrow1,\qquad y\leftarrow z .
  \end{array}
  \label{eq:bbht}
\end{equation}
For any $\lambda\in(1,4/3)$, \citet{boyer1998bbht} show that a search with unknown $M$ requires
$O(\sqrt{N/M})$ Grover iterations in expectation. Across the successive threshold updates,
\citet{durr1996} bound the expected total number of Grover iterations $Q$ needed to reach the minimum by
\begin{equation}
  \mathbb E[Q]\le\tfrac{45}{4}\sqrt N .
  \label{eq:dhbound}
\end{equation}
This bound characterizes the expected search effort, but does not by itself specify a practical
termination criterion; the criteria used in this paper are introduced in Section~\ref{sec:selector}.

\subsection{The lookup oracle and the short-horizon score}
\label{sec:score}

Grover adaptive search uses a threshold oracle to distinguish improving candidates from the remaining
search states \citep{gilliam2021gas}. For the learning-rate search, this paper uses the held-out
evaluation loss as that criterion: each profile $z$ receives a score $f(z)$, obtained by training from
the current checkpoint under that profile. Given the current threshold profile $y$, profiles satisfying
$f(z)<f(y)$ are the improving candidates and are phase-marked by the oracle for subsequent Grover
amplification.

Evaluating this score coherently would require the transformer-training dynamics and the held-out loss
evaluation to be represented inside the quantum computation, which is beyond the implementation studied
here. The experiments of Section~\ref{sec:res-criteria} and Appendix~\ref{app:provenance} therefore compute the
candidate scores classically and store them in a second
table, $\Tscore$, with one score corresponding to each row of $\Tprof$. Thus, $\Tprof$ specifies the
profiles being searched, whereas $\Tscore$ supplies the scores used by the simulated oracle.

Using $\Tscore$, the simulated lookup oracle $O_y$ applies the required threshold marking
\[
  O_y\,|z\rangle=\begin{cases}-\,|z\rangle, & f(z)<f(y),\\ \phantom{-}\,|z\rangle, & f(z)\ge f(y).\end{cases}
\]
The lookup therefore preserves the marking action required by the search while replacing coherent score
evaluation with access to the classically computed scores. Figure~\ref{fig:oracle} contrasts these two
paths: the intended coherent oracle evaluates $f(z)$ on the device, whereas the experiments obtain the
same threshold decision from $\Tscore$.

\paragraph{Oracle convention.}
The lookup experiments use D\"urr--H\o yer minimum finding \citep{durr1996} with the search dynamics of
Eqs.~\eqref{eq:grover}--\eqref{eq:bbht} and Figure~\ref{fig:round}, with candidate losses supplied by
$\Tscore$. The quantum search acts on profile indices, while classical replay constructs $f(z)$; because
the score depends on replay length, $\Tscore$ is defined at a replay horizon $h$.

\paragraph{The replay horizon.}
At checkpoint $t$, $\Tscore$ is constructed by replaying each candidate profile $z$ for $h$ optimizer
steps from the same saved model and optimizer state, using the same sequence of training batches. Let
$L_{t+h}(z)$ denote the resulting held-out evaluation loss. The checkpoint-local score at horizon $h$ is
\[
  f_t^{(h)}(z):=L_{t+h}(z).
\]
The score has the form of a bilevel hyperparameter objective: an outer loss evaluated on the model that a
fixed number of inner optimization steps produces \citep{franceschi2018bilevel}. The inner dynamics are the
$h$ replayed optimizer steps from the checkpoint state, and the outer loss is the held-out evaluation loss.

The horizon was fixed by measuring how well shorter replays preserve the candidate order of a $200$-step
replay (Appendix~\ref{app:horizon}): at checkpoints $200$ and $600$, measured before any $\Tscore$ of this
paper was built, $h=25$ is the shortest horizon with the highest rank agreement, whereas the one-step
ranking is unstable across checkpoints, in line with the bias of short-horizon objectives toward small step
sizes \citep{wu2018shorthorizon} and the long-run loss of one-step gains \citep{roulet2024edge}.
This
paper then applies the held-out loss after $h=25$ replayed steps as the score,
\begin{equation}
  f(z)=f_t^{(25)}(z)=L_{t+25}(z).
  \label{eq:score}
\end{equation}

\paragraph{Selection target.}
The target of a selection at checkpoint $t$ is the minimizer of that checkpoint's own $\Tscore$, computed using
only measurements up to $t+25$:
\begin{equation}
  z^*_t=\arg\min_{z\in\Tprof}f_t^{(25)}(z).
  \label{eq:argmin}
\end{equation}
A search that returns the profile $\hat z_t$ instead of $z^*_t$ has the regret $f(\hat z_t)-f(z^*_t)\ge0$, zero when the
minimizer is returned.

\subsection{Termination criteria}
\label{sec:selector}
Because $M$, the number of profiles better than the current threshold, is unknown, a failed round does not reveal whether a
better profile remains.
\citet{durr1996} therefore end the procedure by a global time-out, and \citet{boyer1998bbht} handle
the case $M=0$ by an $O(\sqrt N)$ time-out. Four termination criteria are compared in this paper, in
two families: fixed-budget criteria, and confidence-based criteria that read the failed rounds as
evidence.
Section~\ref{sec:res-criteria} compares them on the measured score tables to choose the criterion of the closed-loop searches.

\paragraph{Fixed-budget termination.}
During a selection, each round adds its $j$ Grover iterations to the running total, and each success opens
a new threshold visit; both counts are recorded.
Criterion A is the global time-out of \citet{durr1996}, a limit on the effort of the whole run. Let $Q$ be
the Grover iterations spent so far and $V$ the threshold visits so far, the initial one included, with $\lg$
the base-2 logarithm:
\begin{equation}
  \text{A:}\quad \text{stop when } Q+\big(V\lg N\big)^{**}\ \ge\ 22.5\sqrt N+1.4\lg^2 N,\ \text{ return } y .
  \label{eq:critA}
\end{equation}
Criterion B instead limits one threshold visit: when the Grover iterations of the current visit reach the budget that the proof
of Theorem~3 in \citet{boyer1998bbht} assigns to one exponential search, without a better profile found, the run stops:
\begin{equation}
  \text{B:}\quad \text{stop when } \textstyle\sum_{\mathrm{visit}} j\ \ge\ \tfrac12\tfrac{\lambda}{\lambda-1}\sqrt N
  \;=\;3\sqrt N\ \text{ at } \lambda=\tfrac65,\ \text{ return } y .
  \label{eq:critB}
\end{equation}
\paragraph{Derivation note.}
A double asterisk marks a paper-specific extension of a cited bound; the corresponding derivations are
given in Appendix~\ref{app:critA}.
\paragraph{Confidence-based termination.}
The second family treats failed rounds as evidence and stops when they bound (criterion C) or estimate
(criterion D) the chance that a better profile remains below a confidence level $\delta$. Both are
computed classically from the observed rounds.

Criterion C counts the consecutive failed rounds of one visit once the range $m_{\mathrm{BBHT}}$ has saturated, where
\citet[Lemma~2]{boyer1998bbht} guarantee a success probability of at least $\tfrac14$ whenever $M\ge1$; if a better profile
remains, $k$ such failures occur with probability at most $(3/4)^k$, and the count restarts at every threshold update:
\begin{equation}
  \text{C:}\quad (3/4)^k\le\delta\ \Longrightarrow\ k_{\mathrm{stop}}(\delta)=\big\lceil\ln\delta/\ln(3/4)\big\rceil^{**}
  \;=\;25\ \text{ at } \delta=10^{-3} .
  \label{eq:critC}
\end{equation}

Criterion D makes the confidence state-dependent. A success updates the possible rank of the threshold, so
the next failure is interpreted from that new state rather than counted equally with earlier failures. The
resulting distribution $\pi(r)$ over the rank $r$ of the current threshold within $\Tscore$, uniform at the
start, $\pi_0(r)=1/N$, is updated after every round of a visit from the distribution $\pi_u$ held at its
start, and the search stops when the probability that a better profile remains falls below $\delta$,
$\sum_{r\ge2}\pi(r)\le\delta$:
\begin{equation}
  \text{D}^{**}\!:\quad
  \begin{array}{ll}
    \text{failure with } j: & \pi(r)\to\pi(r)\,[1-p_j(r-1)]\,/\,c_\pi,\\[4pt]
    \text{success:} & \pi_{u+1}(r)=\dfrac{1}{c_\pi}\displaystyle\sum_{s>r}\frac{\pi_u(s)}{s-1}.
  \end{array}
  \label{eq:critD}
\end{equation}
$c_\pi$ normalizes $\pi$ to total mass one after each update, and the next visit starts from $\pi_{u+1}$.

\paragraph{Rounds and queries.}
The four termination criteria differ mainly in how long the search continues after the improving part
of the run. The total Grover-iteration count of a selection under criterion $x\in\{\mathrm A,\mathrm B,\mathrm C,\mathrm D\}$
therefore splits into the iterations that improve the threshold and those spent in the terminal visit before stopping:
\begin{equation}
  R^{(x)}=\#\text{rounds}=\#\text{preparations}=\#\text{checks},\qquad
  Q^{(x)}=\textstyle\sum j=Q_{\mathrm{imp}}^{(x)}+Q_{\mathrm{stop}}^{(x)}.
  \label{eq:counts}
\end{equation}
Here the improvement effort $Q_{\mathrm{imp}}^{(x)}$ counts the iterations of visits that find a better threshold and the
stopping effort $Q_{\mathrm{stop}}^{(x)}$ those of the final visit before criterion $x$ stops; each round has one state
preparation, one measurement and one classical check.
For criteria A--D, rounds and Grover iterations summarize search effort, while timing estimates and detailed cost
accounting are given in Appendix~\ref{app:cost}.

\subsection{Interaction-Guided Schedule Oracle}
\label{sec:igso}
\begin{figure}[!htb]
  \centering
  \includegraphics{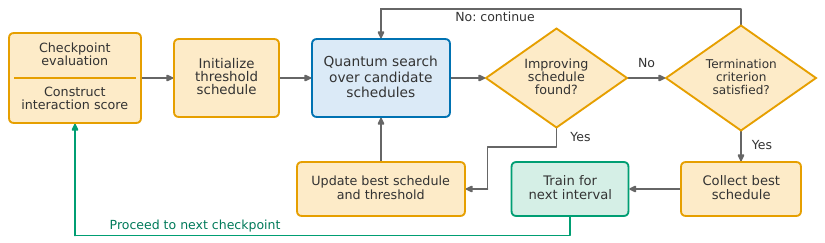}
  \caption{Hybrid loop with the Interaction-Guided Schedule Oracle: classical measurements construct the interaction score,
  quantum search selects a schedule, and training applies it.}
  \label{fig:igso}
\end{figure}

After the termination criterion is chosen from the lookup-oracle data, the Interaction-Guided Schedule Oracle is applied to
the same pipeline without a precomputed score table and works in two steps inside one circuit: it first scores all joint
schedules at once in superposition, adding up the main and interaction effects of the layer groups that apply to each
schedule, and then compares every score with that of the current threshold schedule and marks the schedules that beat it;
Grover amplification increases the chance that the measured schedule is an improving candidate. Figure~\ref{fig:igso}
connects the classical loss measurements with the quantum schedule selection and the subsequent training interval.

\paragraph{Checkpoint evaluation and the interaction score.}
Following the intuition of main effects and interaction effects, a schedule is assumed to change the loss through the effect
of each group alone and through interactions between groups. Pair and triple interactions are kept in this first model;
further terms can be added. In units of the uniform rate, a schedule $z=(\eta_1,\dots,\eta_G)$ of Eq.~\eqref{eq:profile}
takes $\eta_g\in\{0.5,0.7,1.0,1.4\}$, and the uniform-rate baseline $z_{\mathrm{ref}}$ has every $\eta_g=1.0$. For
comparison, the checkpoint evaluation moves the rates of one, two or three groups away from the baseline and takes one trial
optimizer step; the gain is the drop in held-out loss against the same step at the baseline. The main effect $Y_t(g;\eta)$,
the single-group gain, is the gain of moving group $g$ alone to the rate $\eta$, measured at $0.5$, $0.7$ and $1.4$. Pairs and
triples are measured only at $0.5$ and $1.4$, a choice that reduces the measurement and computation requirements; their
interactions $\nu_{t,gg'}$ and $\nu_{t,gg'g''}$ are derived from the joint gains by removing the lower-order effects. An
interaction term with a group at $0.7$ shares the term derived at $0.5$, written $\bar\eta_g=0.5$ when $\eta_g=0.7$ and
$\bar\eta_g=\eta_g$ otherwise. The interaction score of a schedule is
\begin{equation}
  \widehat Y_t(z)=\underbrace{\sum_{g=1}^{G}Y_t(g;\eta_g)}_{\text{single-group gains}}
  +\underbrace{\sum_{g<g'}\nu_{t,gg'}(\bar\eta_g,\bar\eta_{g'})}_{\text{pair interactions}}
  +\underbrace{\sum_{g<g'<g''}\nu_{t,gg'g''}(\bar\eta_g,\bar\eta_{g'},\bar\eta_{g''})}_{\text{triple interactions}} .
  \label{eq:gainmodel}
\end{equation}
Compiled into the oracle circuit as integer constants, the terms of Eq.~\eqref{eq:gainmodel} let the oracle score all
$4{,}096$ schedules in superposition. The interaction score $\widehat Y_t(z)$ plays a role analogous to the kernel of a
support vector machine: its terms can change while the quantum scoring and marking structure is retained.
Appendix~\ref{app:igso} gives the evaluation protocol, interaction derivation and integer score.

\paragraph{Quantum search over candidate schedules.}
Each group's rate takes two qubits, so $q_{\mathrm{srch}}=12$ search qubits hold all $N=K^G=4{,}096$ schedules of $K=4$ rates
over $G=6$ groups. With $\Phi_t(z)$ the integer form of $\widehat Y_t(z)$ and $y$ the current threshold schedule, the oracle
marks every schedule that scores above the threshold:
\begin{equation}
  O_y\,|z\rangle=(-1)^{\mathbf 1[\Phi_t(z)>\Phi_t(y)]}\,|z\rangle .
  \label{eq:igmark}
\end{equation}
The search starts with the uniform-rate baseline $z_{\mathrm{ref}}$, whose score is zero, and retains each improving schedule
it samples as the new threshold. When termination criterion D is satisfied, it returns the best schedule retained, $\hat z_t$,
which sets the group rates of the next $200$ training steps, and the next checkpoint repeats the checkpoint evaluation.

\subsection{Simulation, hardware, and full-run validation}
\label{sec:setup}

The tables of $N=81$, $128$, and $256$ profiles, the noise-free simulation of the search and the trial design
are given in Appendix~\ref{app:setup}, and a $16$-profile instance of the checkpoint-$200$ table is run on an IBM Heron~r2
device to check the action of the oracle and the diffuser (validation and clocks in
Appendix~\ref{app:provenance}). A search returns its threshold row as the selected profile of its
checkpoint, the estimate of $z^*_t$; over the $n_{\mathrm{sel}}$ selection instants $t_1<\dots<t_{n_{\mathrm{sel}}}$
the minimizers form the series $Z^*=(z^*_{t_1},\dots,z^*_{t_{n_{\mathrm{sel}}}})$ and the selected profiles $\hat Z$.
Whether a selected series helps the fine-tuning is checked by training under it: a full run
$\mathrm{FR}(Z^*,\text{seed})$ trains the $n_{\mathrm{tr}}$ steps under the uniform-rate baseline
$z_{\mathrm{ref}}$ until the first selection instant and under each minimizer until the next,
\begin{equation}
  \mathrm{FR}(Z^*,\text{seed}):\qquad
  z(t)=\begin{cases}
    z_{\mathrm{ref}}, & t<t_1,\\
    z^*_{t_i}, & t_i\le t<t_{i+1},\quad i=1,\dots,n_{\mathrm{sel}},
  \end{cases}
  \qquad t_{n_{\mathrm{sel}}+1}=n_{\mathrm{tr}},
  \label{eq:fullrun}
\end{equation}
and the trained model is compared with the run under $z_{\mathrm{ref}}$ by its held-out loss; the executed
runs are recorded in Appendix~\ref{app:fr}. A closed-loop run of Section~\ref{sec:igso} is Eq.~\eqref{eq:fullrun} with
$\hat z_{t_i}$ in place of $z^*_{t_i}$, each search simulated noise-free.

\ifibmloop
Section~\ref{sec:res-ibmloop} repeats the closed loop with every search sampled on an IBM superconducting device, using the
full circuit of Section~\ref{sec:igso}; Appendix~\ref{app:ibm-jlimit} gives the gate limits of the device and the bound they
set on the iteration count.
\fi

\section{Results}
\label{sec:results}

\subsection{Selecting a termination criterion from measured score tables}
\label{sec:res-criteria}
\label{sec:res-scaling}

Hardware validation of the lookup oracle and diffuser on a $16$-profile instance is reported in Appendix~\ref{app:provenance}.
The measured $h=25$ score tables are used to compare the stopping effort and selection quality of criteria A--D and guide the
choice of criterion for the subsequent closed-loop searches.

Table~\ref{tab:criteria} compares the four termination criteria on the $N=81$ tables, pooled over the five checkpoints. All
four spend the same improvement effort and reach comparable selection quality, with a mean regret below the replay drift of
order $10^{-4}$ (Appendix~\ref{app:setup}). Most Grover iterations go into stopping: $76\%$ (D) to $96\%$ (A) of them come
after the last improvement, and criterion A uses about $8.7$ times the stopping effort $Q_{\mathrm{stop}}^{(x)}$ of criterion D.
Criteria B and D are therefore the practical choices. B caps the iterations of each visit with a fixed budget that needs only
$N$; D reads the failed rounds of the visit through the classical confidence update and needs the least stopping effort. This
paper selects criterion D for the closed-loop searches; detailed data are in Appendices~\ref{app:scaling}
and~\ref{app:cost}.

\begin{table}[!h]
  \centering
  \caption{Search effort and selection quality under four termination criteria ($N=81$).}
  \label{tab:criteria}
  \footnotesize
  {\setlength{\tabcolsep}{4.8pt}\begin{tabular}{lccccc}
\toprule
criterion & rounds $R^{(x)}$ & $Q_{\mathrm{imp}}^{(x)}$ & $Q_{\mathrm{stop}}^{(x)}$ & P(min.\ found) & regret \\
\midrule
A global time-out & 77.3 & 8.2 $\pm$ 0.1 & 222.2 $\pm$ 0.2 & 1.000 & 0 \\
{\bfseries\boldmath B per-visit budget $3\sqrt{N}$} & {\bfseries\boldmath 29.3} & {\bfseries\boldmath 8.1 $\pm$ 0.1} & {\bfseries\boldmath 29.4 $\pm$ 0.0} & {\bfseries\boldmath 0.999} & {\bfseries\boldmath $1.9\times10^{-6}$} \\
C confidence count & 52.2 & 8.2 $\pm$ 0.1 & 121.3 $\pm$ 0.2 & 1.000 & 0 \\
{\bfseries\boldmath D state-dependent stop} & {\bfseries\boldmath 28.3} & {\bfseries\boldmath 8.1 $\pm$ 0.1} & {\bfseries\boldmath 25.6 $\pm$ 0.1} & {\bfseries\boldmath 0.999} & {\bfseries\boldmath $1.4\times10^{-6}$} \\
\bottomrule
\end{tabular}}

  \par\vspace{4pt}
  \begin{minipage}{\textwidth}\footnotesize
  Criteria B and D stop after fewer search rounds while maintaining comparable selection quality. $5{,}000$ simulated trials
  per criterion, $\delta=10^{-3}$ for C and D; full table and setting in Appendix~\ref{app:table1}.
  \end{minipage}
\end{table}

\subsection{Closed-loop training with a simulated Interaction-Guided Schedule Oracle}
\label{sec:res-closedloop}
The simulated Interaction-Guided Schedule Oracle drove two closed-loop fine-tunes by selecting joint group-rate schedules at
each of five checkpoints. Each selected schedule was applied to the next training interval, and both resulting models achieved
benchmark accuracies within the range observed across five uniform-rate seeds. The two runs supply different interaction
coefficients to the same Schedule Oracle structure: one takes the checkpoint evaluation on a single training batch, the other
averages it over five batches. The single-batch and the averaged evaluation select different schedules
(Appendix~\ref{app:igso}); comparing the performance of the two is a direction for future research.
Table~\ref{tab:closedloop} compares the runs with the uniform-rate run of the same seed and start, the
range over five uniform-rate seeds and the base model. On every benchmark the two oracle runs differ by at most $0.5$
percentage points, the largest difference from the uniform-rate run is $0.7$ points (single-batch, MMLU), and the
single-batch SciQ score lies at the lower end of the five-seed range. Comparable observed scores from one seed per run do not
establish equivalence, better training or a quantum speedup: the searches are noise-free statevector simulations, and no error
bar exists for the loss of a single run.

\begin{table}[!htb]
  \centering
  \caption{Held-out loss and benchmark accuracy of the closed-loop fine-tunes, seed $2$.}
  \label{tab:closedloop}
  \footnotesize
  \resizebox{\textwidth}{!}{%
  \begin{tabular}{l ccccc cccc}
    \toprule
    & \multicolumn{5}{c}{held-out loss at step} & \multicolumn{4}{c}{accuracy (\%)} \\
    \cmidrule(lr){2-6}\cmidrule(lr){7-10}
    run & $400$ & $600$ & $800$ & $1000$ & $1221$ & SciQ & PubMedQA & ARC-Easy & MMLU \\
    \midrule
    oracle, single batch            & $2.6663$ & $2.6666$ & $2.6710$ & $2.6753$ & $2.6827$ & $89.4$ & $62.1$ & $62.5$ & $38.4$ \\
    oracle, five-batch average      & $2.6661$ & $2.6630$ & $2.6614$ & $2.6682$ & $2.6758$ & $89.6$ & $61.6$ & $62.9$ & $38.7$ \\
    \ifibmloop
    oracle, sampled on IBM hardware & $2.6696$ & $2.6654$ & $2.6645$ & $2.6731$ & $2.6806$ & $89.5$ & $61.3$ & $62.5$ & $38.8$ \\
    \fi
    uniform rate, same seed         & $2.6670$ & $2.6639$ & $2.6673$ & $2.6715$ & $2.6795$ & $89.8$ & $61.6$ & $62.6$ & $39.1$ \\
    uniform rate, five seeds (range) & & & & & & $89.4$--$90.6$ & $60.7$--$64.7$ & $61.3$--$63.3$ & $37.9$--$41.1$ \\
    base model, no fine-tuning      & & & & & & $89.8$ & $59.6$ & $61.7$ & $37.8$ \\
    \bottomrule
  \end{tabular}}
  \par\vspace{4pt}
  \begin{minipage}{\textwidth}\footnotesize
  All runs start from the same checkpoint at step $200$ (held-out loss $2.678$). Held-out loss on $64$ evaluation
  sequences; zero-shot accuracy in percent, standard errors about $1.0$, $1.5$, $1.0$ and $1.4$ points.
  \end{minipage}
\end{table}

\ifibmloop
\subsection{Closed-loop training with the Interaction-Guided Schedule Oracle on IBM hardware}
\label{sec:res-ibmloop}
The loop of Section~\ref{sec:res-closedloop} was repeated with every search sampled on IBM hardware. The five searches of
Table~\ref{tab:ibmloop} ran on three $156$-qubit IBM Heron devices, with one shot per circuit and no simulator sample in any
decision; the largest circuit had $4{,}284{,}266$ two-qubit gates. As on the score tables of Section~\ref{sec:res-criteria},
most Grover iterations went into stopping: $1{,}207$ of the $1{,}381$ ($87\%$) came after the last improvement of each search.
All four benchmark accuracies of the trained model lie within the five-seed range (Table~\ref{tab:closedloop}), under the
single-seed qualification of Section~\ref{sec:res-closedloop}. Appendices~\ref{app:ibm-jlimit} and~\ref{app:ibmloop} give the
device limits, the threshold paths and the IBM turnaround.

\begin{table}[!htb]
  \centering
  \caption{Hardware-sampled searches at training checkpoints.}
  \label{tab:ibmloop}
  \footnotesize
  {\setlength{\tabcolsep}{4pt}\begin{tabular}{r l l rrrrr}
\toprule
checkpoint & device & returned schedule & $\Phi$ & $R$ & $Q$ & $U$ & largest $j$ \\
\midrule
$200$ & \texttt{ibm\_fez}, \texttt{ibm\_marrakesh} & (1.4, 0.7, 1.4, 1.4, 0.5, 1.4) & $42$ & $43$ & $235$ & $6$ & $54$ \\
$400$ & \texttt{ibm\_marrakesh} & (0.7, 0.5, 1.4, 0.7, 1.4, 0.7) & $64$ & $35$ & $262$ & $3$ & $56$ \\
$600$ & \texttt{ibm\_kingston} & (1.4, 1.4, 0.5, 0.5, 0.7, 0.5) & $67$ & $48$ & $307$ & $3$ & $45$ \\
$800$ & \texttt{ibm\_kingston} & (0.5, 1.4, 1.0, 1.4, 0.7, 1.0) & $18$ & $55$ & $298$ & $3$ & $60$ \\
$1000$ & \texttt{ibm\_marrakesh} & (1.0, 0.5, 0.7, 0.7, 1.4, 1.4) & $5$ & $49$ & $279$ & $2$ & $60$ \\
\bottomrule
\end{tabular}
}
  \par\vspace{4pt}
  \begin{minipage}{\textwidth}\footnotesize
  One search per checkpoint. Returned schedule from the first layer group to the last; $\Phi$, final integer interaction
  score; $R$, rounds; $Q$, Grover iterations; $U$, threshold updates; largest $j$, the most Grover iterations in one circuit.
  \end{minipage}
\end{table}
\fi

\section{Summary, Discussion and Limitations}
\label{sec:summary}
Quantum minimum finding was joined to language-model fine-tuning in two parts: a study of termination criteria, and the
Interaction-Guided Schedule Oracle. On a lookup oracle over measured score tables, four termination criteria select with
comparable quality and differ in stopping effort. The Interaction-Guided Schedule Oracle scores all joint schedules in
superposition without a precomputed score table; its selected schedules drove \ifibmloop three closed-loop fine-tunes, one
sampled on IBM hardware,\else two closed-loop fine-tunes\fi{} with benchmark accuracies within the range of uniform-rate
seeds.\footnote{Each closed-loop fine-tune used a single seed, and the scores of unmeasured schedules are modeled.
Establishing training benefits and computational advantage requires further evaluation; Appendix~\ref{app:igso} gives the
modeling and simulation details.} Together, these studies demonstrate how classical checkpoint measurements can support
quantum selection of joint learning-rate schedules within a language-model training loop.

Three directions follow. First, the interaction score can be exchanged while the quantum search is kept. The two simulated
runs used single-batch and five-batch checkpoint evaluations to supply different interaction coefficients to the same quantum
scoring and marking structure (Appendix~\ref{app:igso}). As with the kernel of a support vector machine
\citep{cortes1995svn}, studies of dynamic learning-rate control can supply coefficients for other groupings, rate menus,
models or training phases. Measurements can also be added where evaluations disagree, as in active learning
\citep{cohn1996active} or the reweighting of boosting \citep{freund1997boosting}. The score compiled into the circuit offers
an interface for quantum neural-network research.

Second, quantum selection can refine a baseline schedule $z_{\mathrm{base}}$ proposed by a classical adaptive method, such as
population-based training \citep{jaderberg2017pbt} or hypergradient descent \citep{baydin2018hypergradient}. With the
checkpoint evaluation taken around $z_{\mathrm{base}}$, the interaction score keeps the form of Eq.~\eqref{eq:gainmodel}:
\begin{equation}
  \widehat Y_t(z\mid z_{\mathrm{base}})
  =\underbrace{\sum_{g=1}^{G}Y_t(g;\eta_g)}_{\text{single-group gains}}
  +\underbrace{\sum_{g<g'}\nu_{t,gg'}(\eta_g,\eta_{g'})}_{\text{pair interactions}}
  +\underbrace{\sum_{g<g'<g''}\nu_{t,gg'g''}(\eta_g,\eta_{g'},\eta_{g''})}_{\text{triple interactions}},
  \label{eq:genbase}
\end{equation}
where every gain and interaction is measured around $z_{\mathrm{base}}$ and vanishes when one of its groups stays at its
baseline rate. The baseline scores zero and is the first threshold, so the search looks for joint refinements beyond it. The
present runs use $z_{\mathrm{base}}=z_{\mathrm{ref}}$; other baselines will be studied in future research
(Appendix~\ref{app:igso}).

Third, stopping-policy design is a route to quantum-search efficiency. Most Grover iterations came after the last improvement
\ifibmloop in both the IBM hardware searches and the lookup-oracle study\else in the lookup-oracle study\fi. In that study,
criterion A used about $8.7$ times the stopping effort of criterion D. Criteria B and D are practical choices, and D is used
here. Deciding when enough search has been done can avoid unnecessary rounds and use quantum resources more effectively.

This paper presents an early-stage architecture that integrates quantum learning-rate selection with classical machine
learning. Any practical benefit will depend on advances in quantum hardware and further evaluation. Developing the algorithms
and interfaces now provides a foundation for future research in machine learning and medical imaging.

\bibliographystyle{iclr2027_conference}
\bibliography{references}

\appendix
\setcounter{figure}{0}\renewcommand{\thefigure}{A\arabic{figure}}
\setcounter{table}{0}\renewcommand{\thetable}{A\arabic{table}}
\raggedbottom

\clearpage
\section{Protocols of the replay-horizon study}
\label{app:horizon}
To determine the replay horizon, the study measures how well a shorter replay preserves the candidate order obtained
after a longer one. Let $h_{\max}$ be the longest horizon replayed, $\mathcal A$ a measured subset of the profiles
of $\Tprof$, and $\rho_{\mathrm S}$ the Spearman rank correlation. At checkpoint $t$, the rank agreement of
horizon $h$ with $h_{\max}$ is
\begin{equation}
  \tau_{t,\mathcal A}(h)=\rho_{\mathrm S}\Big(\big(f_t^{(h)}(z)\big)_{z\in\mathcal A},\;\big(f_t^{(h_{\max})}(z)\big)_{z\in\mathcal A}\Big),
  \qquad h_{\max}=200 .
  \label{eq:tau}
\end{equation}
Table~\ref{tab:horizon} reports $\tau_{t,\mathcal A_{16}}(h)$ for $h\in\{1,10,25,50,100,150\}$ at checkpoints
$t\in\{200,600,1000\}$, on the same measured subset $\mathcal A_{16}$ of $16$ profiles at every horizon and
checkpoint.

The one-step ranking is unstable across the three checkpoints, whereas every horizon from $10$ to $150$ agrees
far more strongly with the $200$-step order. The literature reports the same disagreement between one step and
the long run: short-horizon meta-objectives are biased toward small step sizes \citep{wu2018shorthorizon},
one-step gains are overtaken by a constant rate in the long term \citep{roulet2024edge}, and a schedule search
that probes each candidate over a short window fixes that window as a design constant \citep{jin2021autolrs}.

At checkpoints $200$ and $600$, measured before any $\Tscore$ of this paper was built, $h=25$ is the shortest
horizon at which $\tau_{t,\mathcal A_{16}}(h)$ reaches its maximum; the later replay at checkpoint $1000$ agrees
with the $200$-step order above $0.97$ at every horizon from $10$ to $150$, and the setting was kept. %
Table~\ref{tab:horizon} was measured on one run under the uniform profile, on the same $16$ candidate profiles at
every horizon and checkpoint. For $h\ge10$, each profile was replayed for $200$ optimizer steps from the saved
model and optimizer state of the checkpoint, on the same sequence of training batches, and the held-out
evaluation loss was recorded at the marks $h\in\{10,25,50,100,150,200\}$; the five entries of a row at $h\ge10$
therefore come from one replay per profile, each compared with the $200$-step mark of the same replay. The $h=1$
entry follows a different protocol: one optimizer step under each profile, scored on the two held-out sequences of
the scoring batch rather than on the held-out evaluation set. Its value enters Table~\ref{tab:horizon} for its rank
only; the same one-step protocol, on a different design of rows, gives the checkpoint evaluation of Section~\ref{sec:igso}.
\begin{table}[H]
  \centering
  \caption{Rank agreement $\tau_{t,\mathcal A_{16}}(h)$ of Eq.~\eqref{eq:tau} at three checkpoints $t$ of a run under
  the uniform profile (one seed), over the same measured subset $\mathcal A_{16}$ of $16$ candidate profiles.}
  \label{tab:horizon}
  \footnotesize
  \begin{tabular}{lrrrrrr}\toprule
  $h$ & $1^{*}$ & $10$ & $25$ & $50$ & $100$ & $150$ \\ \midrule
  $t=200$ & $0.697$ & $0.909$ & $0.994$ & $0.994$ & $0.977$ & $0.991$ \\
  $t=600$ & $0.677$ & $0.965$ & $0.991$ & $0.985$ & $0.974$ & $0.824$ \\
  $t=1000$ & $-0.253$ & $0.991$ & $0.979$ & $0.982$ & $0.977$ & $0.994$ \\
  \bottomrule
  \end{tabular}
  \par\vspace{2pt}
  \parbox{\linewidth}{\footnotesize $^{*}$\,The $h=1$ entry uses a one-step evaluation protocol; the entries at
  $h\ge10$ use the replay protocol defined above. Both protocols are described above.}
\end{table}

\clearpage
\section{Criteria A--D from the cited bounds}
\label{app:critA}
\paragraph{Criterion A.}
Criterion A, Eq.~\eqref{eq:critA}, is the interrupt of the minimum-finding algorithm of \citet{durr1996}
written in the two counts that a run of this paper records. The convention of that paper charges one time step per
Grover iteration of the exponential search and $\lg N$ time steps per execution of the initialization stage,
which prepares the uniform state over the $N$ rows and marks the rows below the current threshold; that
stage runs once per threshold visit, and the observation and the threshold update are not counted. A run
that has spent $Q$ Grover iterations over $V$ threshold visits, the initial one included, has therefore used
$Q+V\lg N$ time steps under that convention, the left side of Eq.~\eqref{eq:critA}.

The right side is the interrupt of \citet{durr1996}, twice the bound of Lemma~2 of that paper on the expected total time before the
threshold holds the minimum, $m_0=\tfrac{45}{4}\sqrt N+\tfrac{7}{10}\lg^2 N$. The first term bounds the expected number of Grover iterations, Eq.~\eqref{eq:dhbound}. The second term is the initialization cost: by Lemma~1 of
the same paper the row of rank $r$ is ever chosen as the threshold with probability $1/r$, so the expected number of
threshold updates is $\sum_{r=2}^{N}1/r=H_N-1$, with $H_N$ the $N$th harmonic number, and the expected
initialization time is at most $(H_N-1)\lg N\le(\ln N)\lg N=(\ln 2)\lg^2 N\le\tfrac{7}{10}\lg^2 N$.
Stopping at $2m_0=22.5\sqrt N+1.4\lg^2 N$ returns the minimum with probability at least $\tfrac12$ (Theorem~1 of
the same paper).

\paragraph{Criterion B.}
Criterion B, Eq.~\eqref{eq:critB}, limits one threshold visit. Within the current visit the iteration counts $j$ accumulate
across rounds; when the visit has spent the budget that the proof of Theorem~3 in \citet{boyer1998bbht} assigns to one
exponential search, $\tfrac12\tfrac{\lambda}{\lambda-1}\sqrt N$, without finding a better profile, the run stops and returns
the threshold. At $\lambda=\tfrac65$ the budget is $3\sqrt N$, and it restarts at every threshold update.

\paragraph{Criterion C.}
Criterion C, Eq.~\eqref{eq:critC}, counts failed rounds after the range of Eq.~\eqref{eq:bbht} has reached
its cap $m_{\mathrm{BBHT}}=\sqrt N$, where $j$ is drawn uniformly from $\{0,\dots,\lceil\sqrt N\rceil-1\}$.
For $M$ marked rows and $j$ uniform on $\{0,\dots,m-1\}$, \citet[Lemma~2]{boyer1998bbht} give the
probability that one round returns a marked row as $P_m=\tfrac12-\sin(4m\theta)/(4m\sin 2\theta)$ with
$\sin^2\theta=M/N$, and show $P_m\ge\tfrac14$ whenever $m\ge1/\sin 2\theta$. Since
$\sin 2\theta=2\sqrt{M(N-M)}/N$, the condition at $m=\sqrt N$ reads $4M(N-M)\ge N$; the product $M(N-M)$
is smallest at $M=1$ and $M=N-1$, where the condition becomes $4(N-1)\ge N$, so it holds for every
$1\le M\le N-1$ once $N\ge2$, and $m=\lceil\sqrt N\rceil\ge\sqrt N$ satisfies it a fortiori. The threshold
row is never marked, so $M\le N-1$ in every visit, and at $M=0$ every round fails. Within one visit the
threshold, and with it $M$, is fixed; each round prepares the uniform state afresh, draws its own $j$ and
measures once, so the saturated rounds of a visit are independent trials with failure probability
$1-P_m\le\tfrac34$ whenever $M\ge1$. The probability that $k$ consecutive saturated rounds all fail while a
better row exists is therefore at most $(3/4)^k$, and $k_{\mathrm{stop}}(\delta)=\lceil\ln\delta/\ln(3/4)\rceil$
is the smallest $k$ with $(3/4)^k\le\delta$: $17$, $25$ and $33$ for $\delta=10^{-2}$, $10^{-3}$ and
$10^{-4}$. The bound holds for every $M\ge1$ and every visit, and it bounds the event that a visit with a
better row present ends without finding one; a probability over $M$ itself requires a prior, which
criterion D supplies. Rounds before saturation do not enter the count, and the count restarts at every
threshold update.
\paragraph{Criterion D.}
Criterion D, Eq.~\eqref{eq:critD}, keeps a distribution over the rank $r$ of the current threshold in
$\Tscore$, rank $1$ being the lowest score, and updates it from the outcome and the iteration count $j$ of
each round. The first threshold is a row chosen uniformly at random, step~1 of \citet{durr1996}, so its
rank is uniform, $\pi_0(r)=1/N$. A threshold of rank $r$ has exactly $r-1$ rows below it, $M=r-1$, so a
round with $j$ iterations fails with probability $1-p_j(r-1)$ by Eq.~\eqref{eq:grover}; a failed round
multiplies $\pi(r)$ by this likelihood, which is Bayes' rule for that observation, and the factor equals
one at $r=1$ because $p_j(0)=0$. After $j$ iterations from the uniform state the $M$ marked rows carry one
common amplitude $\sin((2j+1)\theta)/\sqrt M$ \citep[Eq.~(3)]{boyer1998bbht}, so a successful round from
rank $s$ returns each of the $s-1$ better rows with probability $1/(s-1)$ whatever $j$ was
\citep[proof of Lemma~1]{durr1996}, and the new rank is uniform on $\{1,\dots,s-1\}$. Summing this
transition over the old ranks $s>r$ gives the success map $\pi_{u+1}(r)\propto\sum_{s>r}\pi_u(s)/(s-1)$,
applied to the distribution $\pi_u$ held at the start of the visit, so $\pi_u$ is the rank distribution of
the $u$-th threshold in the descent of \citet{durr1996} from the uniform start and depends on no $j$. The
failed rounds of the current visit multiply $\pi_u$ in turn, and $c_\pi$ restores total mass one after
every update; at a success $c_\pi=\sum_{s\ge2}\pi_u(s)$, the mass on the ranks from which a success is
possible. The mass on ranks $r\ge2$ is the probability, given the number of threshold updates so far and
the failed rounds of the current visit, that at least one row of $\Tscore$ lies below the current
threshold; the run stops when this mass is at most $\delta$ and returns the threshold.

\clearpage
\section{Cost accounting from the cited analysis}
\label{app:cost}
\paragraph{The three clocks.}
Figure~\ref{fig:round} separates each search round into state preparation and measurement, Grover
iterations, and a classical check. Summed over $R^{(x)}$ rounds and $Q^{(x)}$ Grover iterations, one
quantum selection takes
\begin{equation}
  T_{\mathrm{sel}}^{(x)}=Q^{(x)}\,t_{\mathrm{iter}}+R^{(x)}\,t_{\mathrm{setup}}+R^{(x)}\,t_{\mathrm{chk}} .
  \label{eq:time}
\end{equation}
Here $t_{\mathrm{iter}}$ is the time of one Grover iteration, $t_{\mathrm{setup}}$ the device time for one
state preparation and measurement, and $t_{\mathrm{chk}}$ the time of one classical check.

\paragraph{Training totals.}
With the termination criteria defined, the training cost separates into the ordinary training time
$T_C$ of Eq.~\eqref{eq:tc} and the cost of the $n_{\mathrm{sel}}$ profile selections made during the run.
At each selection, exhaustive classical search evaluates all $N$ profiles at a check time $t_{\mathrm{chk}}$
each, costing $N\,t_{\mathrm{chk}}$, whereas quantum minimum finding incurs a selection time
$T_{\mathrm{sel}}^{(x)}$ that depends on the termination criterion $x\in\{\mathrm A,\mathrm B,\mathrm C,\mathrm D\}$.
The two total costs are therefore
\begin{equation}
  T_C^{\mathrm{exh}}=T_C+n_{\mathrm{sel}}\,N\,t_{\mathrm{chk}},\qquad
  T_Q^{(x)}=T_C+n_{\mathrm{sel}}\,T_{\mathrm{sel}}^{(x)} .
  \label{eq:totals}
\end{equation}
Thus, the simulations compare $T_C^{\mathrm{exh}}$ with $T_Q^{(x)}$, with the difference determined by the
per-selection cost $N\,t_{\mathrm{chk}}$ versus $T_{\mathrm{sel}}^{(x)}$.

\paragraph{Cost convention.}
In the simulation, the classical work used to construct $\Tscore$ is not included in the quantum
selection time $T_{\mathrm{sel}}^{(x)}$. The exhaustive classical selection remains $N\,t_{\mathrm{chk}}$,
Eq.~\eqref{eq:totals}, and the two are compared through the ratio $S=N\,t_{\mathrm{chk}}/T_{\mathrm{sel}}^{(x)}$,
priced with equal unit costs or with hardware-informed clocks (below and Appendix~\ref{app:provenance}).

\paragraph{Counts.}
Every round of Eq.~\eqref{eq:bbht} has one state preparation, $j$ Grover iterations, and one measurement
followed by one classical check, so the number of rounds $R^{(x)}$ of Eq.~\eqref{eq:counts} equals the
number of preparations and the number of checks. The four criteria share the improvement phase on common
random numbers (Appendix~\ref{app:setup}), so the visits that improve the threshold are the same under
every $x$ in a trial that reaches the minimum and $Q_{\mathrm{imp}}^{(x)}$ is the same count; a trial that
a criterion ends before the minimum truncates it. The number of threshold updates $U$ is recorded as a
diagnostic and never priced: by Lemma~1 of \citet{durr1996} the element of rank $r$ becomes the threshold
at some point with probability $1/r$, so $\mathbb E[U]=H_N-1$ with $H_N$ the $N$th harmonic number.

\paragraph{The time convention of the cited analysis.}
\citet{durr1996} count one time step per Grover iteration and $\lg N$ steps per initialization of a
threshold visit, and leave uncounted the stage in which the measured row is compared with the threshold;
Criterion~A transcribes that convention into Eq.~\eqref{eq:critA} (Appendix~\ref{app:critA}). The
accounting of Eq.~\eqref{eq:time} prices that comparison, the classical check, at one $t_{\mathrm{chk}}$
per round, and absorbs the initialization, since every state preparation is one of the $R^{(x)}$ rounds.

\paragraph{The two ratios.}
The unit-cost pricing is the convention of query complexity in which the setup, query and checking costs
of a search are counted in one unit \citep{boyer1998bbht,jeffery2013nested}. At $t_{\mathrm{iter}}=t_{\mathrm{setup}}=t_{\mathrm{chk}}$ the right-hand side of Eq.~\eqref{eq:time} is
$(Q^{(x)}+2R^{(x)})\,t_{\mathrm{chk}}$, so
\begin{equation}
  S_{\mathrm{unit}}=\frac{N\,t_{\mathrm{chk}}}{(Q^{(x)}+2R^{(x)})\,t_{\mathrm{chk}}}=\frac{N}{Q^{(x)}+2R^{(x)}},
  \label{eq:sunit}
\end{equation}
a count ratio that holds for any per-row scoring time. Under the hardware-informed pricing the two quantum
terms of Eq.~\eqref{eq:time} are non-negative, so $T_{\mathrm{sel}}^{(x)}\ge R^{(x)}\,t_{\mathrm{chk}}$ and
\begin{equation}
  S_{\mathrm{hw}}=\frac{N\,t_{\mathrm{chk}}}{T_{\mathrm{sel}}^{(x)}}\le\frac{N}{R^{(x)}},
  \label{eq:shw}
\end{equation}
with equality in the limit $Q^{(x)}\,t_{\mathrm{iter}}+R^{(x)}\,t_{\mathrm{setup}}\ll R^{(x)}\,t_{\mathrm{chk}}$;
the term $R^{(x)}\,t_{\mathrm{chk}}$ is the checking cost of \citet{jeffery2013nested}, the cost of testing
whether a returned candidate is marked, which no device clock removes. The hardware-informed pricing follows the literature on the cost of Grover search:
\citet{cade2023grover} quantify Grover speed-ups beyond the asymptotic worst-case analysis;
\citet{babbush2021focus} conclude that a quadratic speedup alone does not give an early fault-tolerant
device a runtime advantage; \citet{amy2016cost} price a quantum computation in units of classical work;
and \citet{hoefler2023hype} argue that only small-data problems with super-quadratic speedups are useful in
practice. \citet{gundlach2025} list the conditions an end-to-end claim would need in deep learning.

\paragraph{Scope of the timing model.}
Eq.~\eqref{eq:time} prices the lookup study, in which one classical check is one replay of $h=25$ optimizer steps. A round of the
Interaction-Guided Schedule Oracle of Section~\ref{sec:igso} also consists of a preparation and measurement, $j$ Grover iterations
and one classical check, but its check is one evaluation of Eq.~\eqref{eq:igscore} for the measured schedule rather than a
replay, so the clocks of Appendix~\ref{app:provenance} do not transfer to it.\ifibmloop{} On hardware the compiled iteration
bodies differ in gate count from body to body and from threshold to threshold (Appendix~\ref{app:ibm-jlimit}), so
$t_{\mathrm{iter}}$ in Eq.~\eqref{eq:time} can only be an effective average. The cost of a hardware-sampled selection further
contains the checkpoint evaluation of Section~\ref{sec:igso}, compilation and upload encoding, queueing, every physical execution,
including diagnostic shots and circuits submitted but not used by the sequential search, and the training itself; these are
accounted per stage in Appendix~\ref{app:ibmloop}.\fi{} No timing comparison is made for the closed-loop runs.

\clearpage
\section{Provenance of the fitted hardware quantities}
\label{app:provenance}
\begin{figure}[!htb]
  \centering
  \includegraphics{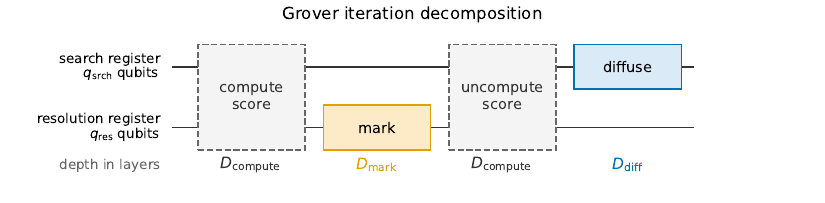}
  \caption{Decomposition of one Grover iteration into score computation, threshold marking, score
  uncomputation, and diffusion. Dashed score blocks denote the declared compute component.}
  \label{fig:iteration-depth}
\end{figure}

\paragraph{Grover-iteration decomposition.}
Figure~\ref{fig:iteration-depth} shows the block decomposition of one Grover iteration. Its scheduled time is
\begin{equation}
  t_{\mathrm{iter}}^{(q_{\mathrm{srch}})}=t_{\mathrm{layer}}^{(q,\,j)}\Big[D_{\mathrm{diff}}^{(q_{\mathrm{srch}},\,1)}+D_{\mathrm{mark}}^{(q_{\mathrm{res}},\,1)}+2\,D_{\mathrm{compute}}\Big].
  \label{eq:hwclock}
\end{equation}
The block timings are estimated from circuits transpiled and scheduled against the calibration of an IBM
Heron~r2 device. The superscript $(q,j)$ records the register size $q$ and Grover-chain setting $j$ used
for the corresponding compile or fit, with $j=1$ denoting a single compiled block. The compilation details and the resulting values follow.
The clocks are anchored by one job on the device: the scheduled duration of a circuit is the preparation with
its readout plus $j$ iteration times,
\begin{equation}
  D(j)=t_0^{(4,3)}+t_{\mathrm{iter}}^{(4,3)}\,j,
  \label{eq:sched}
\end{equation}
fitted for each threshold, and the job's executed time gives $t_{\mathrm{setup}}$.
\paragraph{Superscripts and bands.}
Every quantity of Eq.~\eqref{eq:hwclock} that comes from a compile-and-fit carries a superscript
$(q,j)$: $q$ is the register size the circuit was compiled at ($q_{\mathrm{srch}}$ for the search
register, $q_{\mathrm{res}}$ for the resolution register) and $j$ the largest chain length of the
fit, a straight line over $1,\dots,j$; $j=1$ means one block compiled and its depth read directly.
A quantity without a superscript is a declared choice or a calibration constant of the device.
The layer time is the ratio of the duration slope to the depth slope of the scheduled circuits, a compiler
statistic of the layer mix bounded by the gate durations of the device, so it does not scale with circuit
size. Each fitted depth and the layer time carry a band, the minimum and maximum over three transpiler
seeds and two calibration snapshots around the mean. The setup clock is the executed wall-clock of one round
of the job per shot less its iteration content, so it contains the preparation circuit, the readout and the
device's inter-shot reset delay. Compiling the diffuser and the comparator as one circuit is the cross-check
of the model form; its depth exceeds the sum of the two blocks by the routing the compiler adds. The
break-even of the estimate is the depth $D_{\mathrm{compute}}$ at which the share of the two quantum terms
of Eq.~\eqref{eq:time} in $T_{\mathrm{sel}}^{(x)}$ reaches a given value; below it, $S_{\mathrm{hw}}$ of
Eq.~\eqref{eq:shw} sits at its floor.
\paragraph{The hardware job.}
The job ran on \texttt{ibm\_marrakesh}: seven circuits (the prepared state alone, and $j=1,2,3$ iterations
for two thresholds that mark 4 and 1 of the 16 rows), $4{,}096$ shots each, transpiled at optimization
level~3 and scheduled against the device calibration of the day (median two-qubit gate 68~ns, single-qubit
gate 36~ns, readout $2.68~\mu$s). Eq.~\eqref{eq:sched} is fitted separately for the two thresholds, since
the depth of a lookup oracle depends on which rows it marks; the slope is the iteration time at this size
and the intercept $t_0^{(4,3)}$ the preparation with its readout. The same circuits are scheduled again
against a later calibration to check the slopes, and the job's own execution metric, the QPU time over all
shots, gives the executed cost of one round including reset and repetition delay, from which
$t_{\mathrm{setup}}$ is taken. These four-qubit circuits use a lookup oracle, a diagonal phase over the 16
stored scores, whose depth grows with $N$; the job validates the layer time, the diffuser depth at
$q_{\mathrm{srch}}=4$ and the additivity of iterations, and says nothing about $D_{\mathrm{compute}}$.
\paragraph{Validation of the oracle action.}
Run on the device and in the noise-free simulation, the oracle and the diffuser of these circuits act as designed
on a 16-profile instance of the checkpoint-200 table. With four profiles marked, one iteration raises the
probability of a marked profile from $0.25$ to $0.85$ on the device, against $1.00$ in the simulation. With
the minimizer alone marked, three iterations raise its probability from $0.07$ to $0.63$ on the device,
against $0.96$ in the simulation; the deeper the circuit, the wider the gap, so depth is the quantity a
future circuit has to keep in hand.
\begin{center}
  \footnotesize
  \begin{tabular}{p{1.9cm}p{2.5cm}p{4.4cm}p{3.2cm}}
  \toprule
  quantity & what it is & how obtained & value (band) \\
  \midrule
  $t_{\mathrm{layer}}^{(4..9,3)}$ & layer time & median of $t_{\mathrm{iter}}^{(q,3)}/D_{\mathrm{iter}}^{(q,3)}$ over 34 schedules, $q_{\mathrm{srch}}=4$ to $9$, two oracle families, three seeds, two backends & $33$~ns ($29$--$37$) \\
  $t_{\mathrm{iter}}^{(4,3)}$, $t_0^{(4,3)}$ & one lookup iteration; preparation + readout at $q_{\mathrm{srch}}=4$ & slope and intercept of Eq.~\eqref{eq:sched} over $j=1,2,3$; executed job & $2.56$ / $5.16~\mu$s; $2.7~\mu$s \\
  $D_{\mathrm{diff}}^{(18,1)}$ & diffuser depth, search register & one block compiled at $q_{\mathrm{srch}}=18$; mean over three seeds, two backends & $730$ layers ($706$--$747$) \\
  $D_{\mathrm{mark}}^{(16,1)}$ & comparator depth, resolution register & one block compiled at $q_{\mathrm{res}}=16$; same statistic & $942$ layers ($896$--$969$) \\
  $D_{\mathrm{block}}^{(18,16;\,1)}$ & diffuser and comparator compiled as one circuit & one block, 65 qubits; same statistic; cross-check only & $2{,}057$ layers ($1{,}986$ to $2{,}187$); $68~\mu$s ($66$ to $73$) \\
  $D_{\mathrm{compute}}$ & coherent evaluation of the replay score $f(z)$ & no circuit is built for it & declared; break-even under ``The clocks as measured.'' below \\
  $q_{\mathrm{res}}$ & resolution register width & $\ge\log_2(\text{range}/\sigma)$ with range $4$, $\sigma=0.0017$ & $16$ (derived $\approx12$) \\
  $t_{\mathrm{setup}}$ & one round's executed wall-clock less its iterations & $276.2-6.6~\mu$s from the job metric & $270~\mu$s \\
  $t_{\mathrm{chk}}$ & one classical check & median of $389$ timed $h=25$ replays & $355.5$~s (10th--90th pct.\ $354.4$ to $357.0$) \\
  \bottomrule
  \end{tabular}
\end{center}

\paragraph{The clocks as measured.}
Table~\ref{tab:clocks} summarizes the three clocks of Eq.~\eqref{eq:time}. The layer time of
Eq.~\eqref{eq:hwclock} is $t_{\mathrm{layer}}^{(4..9,3)}=33$~ns, the median of $34$ schedules from
$q_{\mathrm{srch}}=4$ to $9$, stable at $33\pm4$~ns across register sizes and constant in $j$; the
two oracle families of the executed job give $31.8$ and $34.5$~ns. The two known depths at the
design size are $D_{\mathrm{diff}}^{(18,1)}=730$ layers ($706$--$747$) and
$D_{\mathrm{mark}}^{(16,1)}=942$ layers ($896$--$969$), a known part of $1{,}672$ layers, so
\begin{equation*}
  t_{\mathrm{iter}}^{(18)}=55~\mu\mathrm{s}\ (53\text{--}57)+66~\mathrm{ns}\times D_{\mathrm{compute}}.
\end{equation*}
Compiling the two blocks as one circuit gives $2{,}057$ layers and $68~\mu$s, about $20\%$ of
routing depth above the model form, which remains the printed number. The resolution register
follows from a loss range of $4$ and a seed-to-seed standard deviation of $0.0017$ along the
reference run, $\log_2(4/0.0017)\approx12$ bits, and $q_{\mathrm{res}}=16$ is the margin above it.
The executed job gives the slopes of Eq.~\eqref{eq:sched} as $t_{\mathrm{iter}}^{(4,3)}=2.56~\mu$s
for the threshold that marks 4 of the 16 rows and $5.16~\mu$s for the one that marks 1, with
intercepts $t_0^{(4,3)}=2.7$ to $3.1~\mu$s dominated by readout; scheduled again against a later
calibration, the slopes reproduce within about $5\%$. The setup clock is
$t_{\mathrm{setup}}^{(4,3)}=270~\mu$s: the job's $28{,}672$ shots took $7.92$~s of QPU time, $276.2~\mu$s
per shot, less the $6.6~\mu$s of iteration content pooled over its seven circuits; the preparation
circuit's own share is $t_0^{(4,3)}$ and the rest is the device's inter-shot reset delay, a backend
setting taken as independent of the circuit. One classical check costs $t_{\mathrm{chk}}=355.5$~s,
the median of $389$ timed replays (10th to 90th percentile $354.4$ to $357.0$~s); a later campaign of
the same replay gave $336.6$~s, and the slower value is used.
One replay restores the checkpoint with its optimizer state and then runs $h\,t_{\mathrm{tr}}=25\times12.4~\mathrm{s}=310$~s of
optimizer steps. With these clocks and the counts at
$N=2^{18}$ of Appendix~\ref{app:scaling}, the classical checks of one selection take
$R^{(\mathrm D)}\,t_{\mathrm{chk}}=70{,}000$~s, $19.5$~h, and the two quantum terms of Eq.~\eqref{eq:time}
total less than $0.4$~s when $D_{\mathrm{compute}}=0$. The share of the two terms reaches $1\%$ only if $D_{\mathrm{compute}}$
exceeds about $10^6$ layers at the Heron clock, or about $10^4$ layers at a hypothetical
fault-tolerant layer of $7.7~\mu$s, seven cycles of a distance-7 surface code at the $1.1~\mu$s
cycle of \citet{acharya2025willow}; below that, $S_{\mathrm{hw}}$ of Eq.~\eqref{eq:shw} sits at its
floor $N/R^{(x)}$. At the measured sizes $N\le256$ the two quantum terms total less than $0.1$~s
against $R^{(x)}\,t_{\mathrm{chk}}$ of $2.8$ to $4.1$~h, so the choice of clock changes no measured
row; it matters only for the continuation to large $N$.

\begin{table}[!htb]
  \centering
  \caption{The three clocks of one selection, Eq.~\eqref{eq:time}: the value as printed, how it
  was obtained, its source, and a literature comparison. $t_{\mathrm{iter}}$ follows
  Eq.~\eqref{eq:hwclock} with $D_{\mathrm{compute}}$ declared; $t_{\mathrm{setup}}$ is taken from
  the executed job described above; $t_{\mathrm{chk}}$ is a GPU clock. The
  compile-and-fit quantities behind the first row are listed in the provenance table above.}
  \label{tab:clocks}
  {\small
\begin{tabular}{p{0.12\textwidth}p{0.15\textwidth}p{0.28\textwidth}p{0.13\textwidth}p{0.17\textwidth}}
\toprule
clock & value (as printed) & how obtained & source & literature comparison \\
\midrule
$t_{\mathrm{iter}}^{(18)}$, one Grover iteration at the design size & $55\,\mu$s (53--57) $+$ $66$\,ns $\times$ $D_{\mathrm{compute}}$ & Eq.~\eqref{eq:hwclock}: $t_{\mathrm{layer}}=33$\,ns, median of 34 compiled schedules ($q_{\mathrm{srch}}=4$ to 9; the executed $q_{\mathrm{srch}}=4$ job gives 31.8 and 34.5\,ns) $\times$ $[D_{\mathrm{diff}}^{(18,1)}=730$ (706--747) $+\,D_{\mathrm{mark}}^{(16,1)}=942$ (896--969)$]=1,672$ layers (1,602--1,716), bands over 3 seeds $\times$ 2 calibration snapshots; $D_{\mathrm{compute}}$ declared & compiled schedules (this work); App.~part A & IonQ Forte MS two-qubit gate median 672\,$\mu$s vs Heron cz 68\,ns: about $10^4\times$ slower per two-qubit gate \citep{chen2024ionq}; Willow surface-code cycle 1.1\,$\mu$s, a logical layer $\approx d$ cycles \citep{acharya2025willow} \\
$t_{\mathrm{setup}}^{(4,3)}$, one state preparation + measurement & $270\,\mu$s & $t_{\mathrm{shot}}^{(4,3)}-$ pooled iteration content $=276.2-6.6=269.6\,\mu$s $\to270$; $t_{\mathrm{shot}}^{(4,3)}=T_{\mathrm{exec}}/\mathrm{shots}=7.920$\,s$/28,672$, one total over the seven circuits ($j=0\dots3$, both families); circuit share $t_0^{(4,3)}=2.7$--3.1\,$\mu$s, the rest is the backend's inter-shot repetition delay (configured 250\,$\mu$s) & job metrics; backend configuration & The cited AES cost model expresses runtime through circuit depth and cycle time, with no setup term \citep{sarahd2024grovercost} \\
$t_{\mathrm{chk}}$, one classical check & $355.5$\,s ($336.6$\,s later population) & median gap between consecutive $h=25$ replay files (restore $+$ 25 steps $+$ evaluation $+$ save): 389 gaps of one replay session, p10--p90 354.4--357.0\,s, the slower population printed; later replays: 336.6\,s; a GPU clock, never averaged & training-run replay timing (this work) & --- (own workload; Eq.~\eqref{eq:totals} $N\,t_{\mathrm{chk}}$, Eq.~\eqref{eq:time} $R^{(x)}t_{\mathrm{chk}}$) \\
\bottomrule
\end{tabular}}

\end{table}

\clearpage
\section{Experimental and implementation details}
\label{app:setup}
\begin{figure}[!htb]
  \centering
  \includegraphics{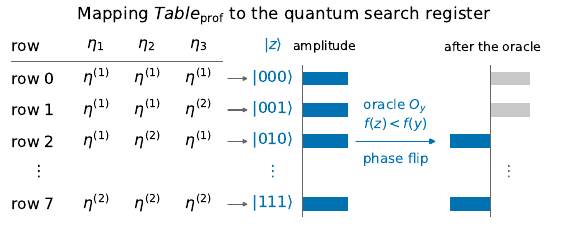}
  \caption{$\Tprof$ mapped to the quantum search register, at an illustrative size ($K=2$ admissible
  rates $\eta_g\in\mathcal E$, $G=3$ groups, $N=8$ rows, $q_{\mathrm{srch}}=3$ qubits). Each row of $\Tprof$ is
  one profile $z$ of Eq.~\eqref{eq:profile}, and its row index written in binary is the basis state
  $|z\rangle$ of the register; every row enters the search with the same amplitude, the state $|\psi_0\rangle$
  of Eq.~\eqref{eq:encode}. The right-hand column shows the amplitudes after the oracle has flipped the
  sign of the selected rows.}
  \label{fig:object}
\end{figure}
\begin{figure}[!htb]
  \centering
  \includegraphics{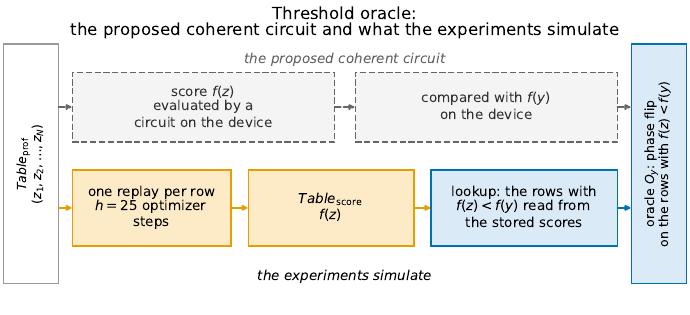}
  \caption{Threshold oracle: the proposed coherent circuit and what the experiments simulate. A quantum
  search over $\Tprof$ needs an oracle that flips the sign of every row whose score lies below the
  threshold, $f(z)<f(y)$. The proposed circuit evaluates the score $f(z)$ on the device and compares it
  with $f(y)$ there. The experiments simulate this behavior: the scores are computed classically, one
  replay of $h=25$ optimizer steps per row, and stored in $\Tscore$, and a lookup applies the same
  phase flip from the stored values. The searches of Section~\ref{sec:res-criteria} and Appendix~\ref{app:provenance}
  run on this simulated oracle, and the rounds of those searches are the recorded data; the closed-loop
  searches of Section~\ref{sec:res-closedloop} use the oracle of Section~\ref{sec:igso}.}
  \label{fig:oracle}
\end{figure}
\begin{figure}[!htb]
  \centering
  \includegraphics{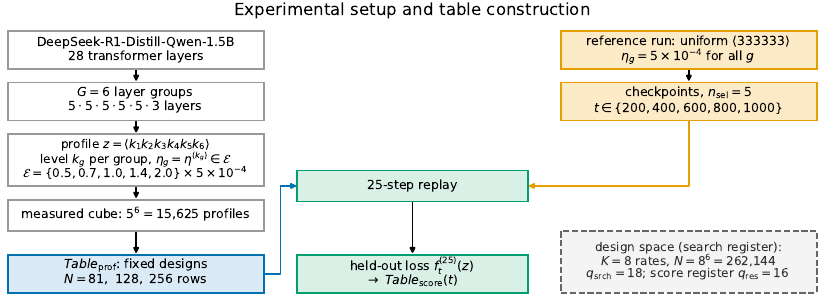}
  \caption{Construction of the score tables: fixed designs of $N=81$, $128$ and $256$ profiles from the
  measured five-level cube, scored at the five checkpoints of the uniform-rate run by the 25-step
  replay of Eq.~\eqref{eq:score}.}
  \label{fig:setup}
\end{figure}
\paragraph{The tables.}
Figure~\ref{fig:setup} summarizes the experimental setup: the reference run and its five checkpoints, the
tables of $N=81$, $128$ and $256$ rows drawn from the measured cube, and the 25-step replay that scores
every row, so $n_{\mathrm{sel}}=5$. The model, training, table-construction and replay details are given
below.

\begin{figure}[!htb]
  \centering
  \includegraphics{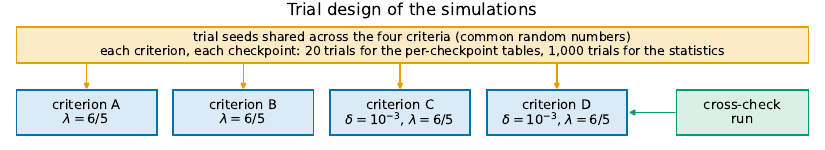}
  \caption{Trial design: the four criteria run on shared seeds at every checkpoint; the cross-check run
  supports criterion D.}
  \label{fig:trials}
\end{figure}
\paragraph{The search.}
The search runs in a noise-free statevector simulation; Figure~\ref{fig:trials} shows the trial design.
The declared-rule cross-check run supports criterion D: validated at the three measured sizes
(Appendix~\ref{app:scaling}), it carries the rule to larger $N$. State preparation, simulation
versions, the oracle implementation, the rank bookkeeping and the cross-check run are given
below.
\paragraph{Model and training configuration.}
The model is DeepSeek-R1-Distill-Qwen-1.5B (28 transformer layers), fine-tuned with LoRA adapters
of rank 16 on every attention and MLP projection, on the peS2o scientific-text corpus, with AdamW
and $16{,}384$ tokens per optimizer step on one NVIDIA RTX~5050 GPU (8~GB).
The layers form $G=6$ groups of five (three in the last). The reference run trains the uniform profile
$\eta_g=5\times10^{-4}$ for all $g$, and its checkpoints at steps $t\in\{200,400,600,800,1000\}$ are
the five tables' origins, so $n_{\mathrm{sel}}=5$.
\paragraph{Profile-table construction.}
The design table of Sections~\ref{sec:object}--\ref{sec:selector} and Appendix~\ref{app:provenance} has $K=8$ admissible rates,
$N=8^6=2^{18}=262{,}144$ rows and $q_{\mathrm{srch}}=18$, with a resolution register of
$q_{\mathrm{res}}=16$ qubits for the score; the measured eta space is the $K=5$ subset
$\mathcal E=\{0.5,\,0.7,\,1.0,\,1.4,\,2.0\}\times5\times10^{-4}$, $5^6=15{,}625$ rows, from which
tables of $N=81$, $128$ and $256$ rows are drawn by fixed designs (anchor profiles, two orthogonal
arrays, a random set, and stride grids); the row lists are part of the data record. A profile is
written by its levels, $z=\langle k_1k_2k_3k_4k_5k_6\rangle$ with group $g$ at rate $\eta^{(k_g)}$; the
uniform-rate baseline $z_{\mathrm{ref}}$ is $\langle333333\rangle$.
\paragraph{Replay and checkpoint protocol.}
At each checkpoint every row is scored by Eq.~\eqref{eq:score}: $25$ optimizer steps replayed from the
frozen checkpoint under the row's profile, with the same batch sequence and a cloned optimizer state
for every row. Replaying the uniform profile itself from the checkpoint at step $200$ reproduces the logged
loss of the reference run exactly at the restart step; over the following $49$ steps the replay drifts by
$1.9\times10^{-4}$ in loss on average, and the evaluation at step $50$ of the replay lands within
$1.7\times10^{-4}$ of the logged value.
On the $h=25$ tables of Figure~\ref{fig:setup}, the resolution of the score is the drift of the replay itself, of
order $10^{-4}$ in loss. Along the run, the score tables of adjacent checkpoints have a rank agreement of $0.919$ on the $81$
profiles (the mean Spearman correlation of the four adjacent pairs), so the candidate order changes slowly as training proceeds.
\paragraph{Quantum simulation configuration.}
The search runs in Qiskit~2.4.2 with Qiskit~Aer~0.17.2 (statevector method), noise-free. For $N=128$
and $256$ the uniform state over the rows is $H^{\otimes q_{\mathrm{srch}}}$ on the
$q_{\mathrm{srch}}=\log_2 N$ qubits of the search register and the diffuser is the standard reflection;
for $N=81$ the uniform state over the 81 valid rows is prepared by a dense Householder unitary with the
matching reflection, so no padding state is ever marked or measured. The oracle is a diagonal phase over
$\Tscore$, thresholded at the score $f(y)$ of the current threshold row. Trial seeds are shared across
criteria (common random numbers), so the four criteria share the visits that improve the threshold and
differ only in the terminal visit, a paired comparison. Ranks are attached after each trial for
evaluation only and are never read by the run. The seed-stability check of Appendix~\ref{app:seeds} uses
four further disjoint seed blocks of $1{,}000$ trials. The declared-rule cross-check run replays
criterion D, with $\delta$ and $\lambda$ unchanged, on ranks alone: the rank $r$ of the initial threshold is
drawn uniformly from $1$ to $N$, a round with $j$ iterations succeeds with the analytic probability of
Eq.~\eqref{eq:grover} at $M=r-1$ rows below the threshold, and a success moves the threshold to a rank drawn
uniformly below the old one; no score table and no circuit enter. It is run at $N=81$, $128$ and $256$
against the simulations (Appendix~\ref{app:scaling}) and then continued to larger $N$.

\clearpage
\section{Recorded outcome of the executed full runs}
\label{app:fr}
Every run of this section is a full run of Eq.~\eqref{eq:fullrun}, $n_{\mathrm{tr}}=1{,}221$ steps under the setup of
Section~\ref{sec:setup}, either under a series $Z^*(\text{table})$ of Section~\ref{sec:setup}, whose members are the
minimizers of one table at the five selection instants, or under one profile held from the first step; the seed fixes the adapter initialization and the batch
order, and $\mathrm{FR}(z,\text{seed})$ denotes the run that holds one profile $z$ from the first step. The series
executed are $Z^*(\text{table }81)=(\langle131111\rangle,\langle111111\rangle,\langle111111\rangle,\langle111111\rangle,\langle111111\rangle)$,
$Z^*(\text{table }128)=(\langle211221\rangle,\langle111112\rangle,\langle111111\rangle,\langle111111\rangle,\langle111112\rangle)$
and $Z^*(\text{table }256)=(\langle112111\rangle,\langle111112\rangle,\langle111111\rangle,\langle111111\rangle,\langle111112\rangle)$;
every member of a series was also trained as a constant profile. The series were measured on the checkpoints of one
uniform run, so in a series run every member after the first acts on a state it was not selected on: the loop is
open, and the tables record an outcome, not a winner.

Each run is read by its held-out evaluation loss $\ell(\mathrm{FR},t_{\mathrm{step}})$, the mean cross-entropy over a
fixed set of $64$ held-out sequences of $1{,}024$ tokens, written at $t_{\mathrm{step}}\in\{0,50,\dots,1200,1221\}$,
and compared at equal steps with the mean $\bar\ell(t_{\mathrm{step}})$ of the uniform runs
$\mathrm{FR}(\langle333333\rangle,\text{seed})$ over the reference seeds. The statistic is the step-matched excess
loss and its mean over the tail window $\mathcal W=\{800,850,\dots,1200,1221\}$ of the last ten evaluations,
\[
  \Delta(\mathrm{FR},t_{\mathrm{step}})=\ell(\mathrm{FR},t_{\mathrm{step}})-\bar\ell(t_{\mathrm{step}}),\qquad
  \bar\Delta(\mathrm{FR})=\frac{1}{|\mathcal W|}\sum_{t_{\mathrm{step}}\in\mathcal W}\Delta(\mathrm{FR},t_{\mathrm{step}}),
\]
together with $n_-$, the number of evaluations in $\mathcal W$ at which the run lies below the reference.
The reference is the mean of the five uniform runs, seeds $0$ to $4$; each uniform run is scored leave-one-out,
against the mean of the other four, and the five uniform rows give the seed-to-seed scale of $\bar\Delta$. The
uniform run of seed $2$ is the run whose checkpoints are the origin of the tables. The second table compares each
series run with runs
of its own seed, $\ell(\mathrm{FR}_a,t_{\mathrm{step}})-\ell(\mathrm{FR}_b,t_{\mathrm{step}})$ averaged over
$\mathcal W$: against the uniform run of that seed, and against each of its members held from the first step.

The last column of the first table is a consistency check on public question-answering sets, run zero-shot with the
EleutherAI evaluation harness on the saved adapter of each run (SciQ, $1{,}000$ questions; PubMedQA, $1{,}000$;
ARC-Easy, $2{,}376$). A run is flagged when its accuracy on any of the three falls below the lowest accuracy of the
uniform runs by more than two standard errors of that accuracy. This is a screening rule fixed before the runs, not a
hypothesis test at a stated level: the band minimum is itself the lowest of a few noisy scores, every run is checked
on three sets, and the seed-to-seed spread of an accuracy is of the same size as its question-sampling standard error.
A test at a stated level would need of the order of eight seeds per profile on the noisiest set and is left to later
work. No run is flagged.
\begin{center}
  \footnotesize
  \resizebox{\textwidth}{!}{\begin{minipage}{1.25\textwidth}\begin{tabular}{llccc}
\toprule
run & tail-mean excess loss & evaluations below reference & final evaluation loss & benchmark flag \\
\midrule
FR($\langle$333333$\rangle$, seed 0) & $-$0.0011 & 6 of 10 & 2.6842 & no \\
FR($\langle$333333$\rangle$, seed 1) & +0.0006 & 4 of 10 & 2.6798 & no \\
FR($\langle$333333$\rangle$, seed 2) & $-$0.0013 & 8 of 10 & 2.6795 & no \\
FR($\langle$333333$\rangle$, seed 3) & +0.0016 & 2 of 10 & 2.6831 & no \\
FR($\langle$333333$\rangle$, seed 4) & +0.0001 & 6 of 10 & 2.6809 & no \\
FR($\langle$232231$\rangle$, seed 0) & $-$0.0074 & 10 of 10 & 2.6721 & no \\
FR($\langle$212222$\rangle$, seed 0) & $-$0.0130 & 10 of 10 & 2.6634 & no \\
FR($\langle$212222$\rangle$, seed 1) & $-$0.0135 & 10 of 10 & 2.6641 & no \\
FR($\langle$131111$\rangle$, seed 0) & $-$0.0123 & 10 of 10 & 2.6632 & no \\
FR($\langle$131111$\rangle$, seed 1) & $-$0.0140 & 10 of 10 & 2.6626 & no \\
FR($\langle$211221$\rangle$, seed 0) & $-$0.0141 & 10 of 10 & 2.6612 & no \\
FR($\langle$211221$\rangle$, seed 1) & $-$0.0146 & 10 of 10 & 2.6630 & no \\
FR($\langle$112111$\rangle$, seed 0) & $-$0.0147 & 10 of 10 & 2.6593 & no \\
FR($\langle$112111$\rangle$, seed 1) & $-$0.0162 & 10 of 10 & 2.6605 & no \\
FR($\langle$111112$\rangle$, seed 0) & $-$0.0159 & 10 of 10 & 2.6583 & no \\
FR($\langle$111112$\rangle$, seed 1) & $-$0.0172 & 10 of 10 & 2.6593 & no \\
FR($\langle$111111$\rangle$, seed 0) & $-$0.0159 & 10 of 10 & 2.6576 & no \\
FR($\langle$111111$\rangle$, seed 1) & $-$0.0172 & 10 of 10 & 2.6594 & no \\
FR($Z^*$(table 81), seed 0) & $-$0.0162 & 10 of 10 & 2.6588 & no \\
FR($Z^*$(table 81), seed 1) & $-$0.0173 & 10 of 10 & 2.6596 & no \\
FR($Z^*$(table 128), seed 0) & $-$0.0162 & 10 of 10 & 2.6595 & no \\
FR($Z^*$(table 128), seed 1) & $-$0.0168 & 10 of 10 & 2.6604 & no \\
FR($Z^*$(table 256), seed 0) & $-$0.0160 & 10 of 10 & 2.6595 & no \\
FR($Z^*$(table 256), seed 1) & $-$0.0168 & 10 of 10 & 2.6605 & no \\
\bottomrule
\end{tabular}

\begin{tabular}{llcc}
\toprule
series run & compared with & difference & evaluations lower \\
\midrule
FR($Z^*$(table 128), seed 0) & FR($\langle$333333$\rangle$, seed 0) & $-$0.0153 & 10 of 10 \\
FR($Z^*$(table 128), seed 0) & FR($\langle$211221$\rangle$, seed 0) & $-$0.0021 & 10 of 10 \\
FR($Z^*$(table 128), seed 0) & FR($\langle$111112$\rangle$, seed 0) & $-$0.0003 & 5 of 10 \\
FR($Z^*$(table 128), seed 0) & FR($\langle$111111$\rangle$, seed 0) & $-$0.0002 & 5 of 10 \\
FR($Z^*$(table 81), seed 0) & FR($\langle$333333$\rangle$, seed 0) & $-$0.0153 & 10 of 10 \\
FR($Z^*$(table 81), seed 0) & FR($\langle$131111$\rangle$, seed 0) & $-$0.0039 & 10 of 10 \\
FR($Z^*$(table 81), seed 0) & FR($\langle$111111$\rangle$, seed 0) & $-$0.0002 & 6 of 10 \\
FR($Z^*$(table 256), seed 0) & FR($\langle$333333$\rangle$, seed 0) & $-$0.0152 & 10 of 10 \\
FR($Z^*$(table 256), seed 0) & FR($\langle$112111$\rangle$, seed 0) & $-$0.0013 & 9 of 10 \\
FR($Z^*$(table 256), seed 0) & FR($\langle$111112$\rangle$, seed 0) & $-$0.0001 & 5 of 10 \\
FR($Z^*$(table 256), seed 0) & FR($\langle$111111$\rangle$, seed 0) & $-$0.0001 & 5 of 10 \\
FR($Z^*$(table 128), seed 1) & FR($\langle$333333$\rangle$, seed 1) & $-$0.0173 & 10 of 10 \\
FR($Z^*$(table 128), seed 1) & FR($\langle$211221$\rangle$, seed 1) & $-$0.0023 & 10 of 10 \\
FR($Z^*$(table 128), seed 1) & FR($\langle$111112$\rangle$, seed 1) & +0.0004 & 3 of 10 \\
FR($Z^*$(table 128), seed 1) & FR($\langle$111111$\rangle$, seed 1) & +0.0004 & 4 of 10 \\
FR($Z^*$(table 81), seed 1) & FR($\langle$333333$\rangle$, seed 1) & $-$0.0177 & 10 of 10 \\
FR($Z^*$(table 81), seed 1) & FR($\langle$131111$\rangle$, seed 1) & $-$0.0033 & 10 of 10 \\
FR($Z^*$(table 81), seed 1) & FR($\langle$111111$\rangle$, seed 1) & $-$0.0001 & 5 of 10 \\
FR($Z^*$(table 256), seed 1) & FR($\langle$333333$\rangle$, seed 1) & $-$0.0173 & 10 of 10 \\
FR($Z^*$(table 256), seed 1) & FR($\langle$112111$\rangle$, seed 1) & $-$0.0006 & 8 of 10 \\
FR($Z^*$(table 256), seed 1) & FR($\langle$111112$\rangle$, seed 1) & +0.0004 & 3 of 10 \\
FR($Z^*$(table 256), seed 1) & FR($\langle$111111$\rangle$, seed 1) & +0.0004 & 4 of 10 \\
\bottomrule
\end{tabular}
\end{minipage}}
\end{center}

\clearpage
\section{Termination criteria per checkpoint}
\label{app:table1}
Table~\ref{tab:criteria-full} repeats Table~\ref{tab:criteria} at full width, with the two cost ratios of Appendix~\ref{app:cost}.
\begin{table}[!htb]
  \centering
  \caption{Four termination criteria for checkpoint-local D\"urr--H\o yer selection on the $h=25$ tables, $N=81$, pooled over
  five checkpoints, $1{,}000$ Aer trials per checkpoint and criterion, common random numbers. Rounds $R^{(x)}$ count state
  preparations and classical checks alike (Eq.~\eqref{eq:counts}); queries split into the descent work $Q_{\mathrm{imp}}^{(x)}$,
  identical across criteria, and the stopping work $Q_{\mathrm{stop}}^{(x)}$, where the criteria differ. The two speedups are
  $S_{\mathrm{unit}}$ of Eq.~\eqref{eq:sunit} and $S_{\mathrm{hw}}$ of Eq.~\eqref{eq:shw}, the latter drawn with the interim clocks
  $t_{\mathrm{iter}}=1$~ms, $t_{\mathrm{setup}}=40~\mu$s; the estimate of Eq.~\eqref{eq:hwclock} is smaller on both and leaves every
  entry at the floor $N/R^{(x)}$ to two digits.}
  \label{tab:criteria-full}
  \footnotesize
  \resizebox{\textwidth}{!}{\begin{tabular}{lccccccc}
\toprule
criterion & rounds & queries, improve & queries, stop & P(min.\ found) & regret & speedup, unit-cost & speedup, hw-informed \\
\midrule
A DH96 global cutoff & 77.3 & 8.2 $\pm$ 0.1 & 222.2 $\pm$ 0.2 & 1.000 & 0 & 0.21$\times$ & 1.05$\times$ \\
B BBHT cascade budget $3\sqrt{N}$ & 29.3 & 8.1 $\pm$ 0.1 & 29.4 $\pm$ 0.0 & 0.999 & 1.9e-06 & 0.84$\times$ & 2.77$\times$ \\
C BBHT-confidence stop, $\delta=10^{-3}$ & 52.2 & 8.2 $\pm$ 0.1 & 121.3 $\pm$ 0.2 & 1.000 & 0 & 0.35$\times$ & 1.55$\times$ \\
D dynamic stop (N,u,f,$\delta$), $\delta=10^{-3}$ & 28.3 & 8.1 $\pm$ 0.1 & 25.6 $\pm$ 0.1 & 0.999 & 1.4e-06 & 0.90$\times$ & 2.86$\times$ \\
\bottomrule
\end{tabular}
}
\end{table}
Per-checkpoint detail behind Table~\ref{tab:criteria}: four criteria, five checkpoints, 20 Aer
trials each, $\delta=10^{-3}$ for C and D. The minimizer column names the row of
Eq.~\eqref{eq:argmin} at that checkpoint; the threshold-update count is a diagnostic
(Appendix~\ref{app:cost}); the rank percentile is $1$ when the minimum was returned.
\begin{center}
  \scriptsize
  \resizebox{\textwidth}{!}{\begin{tabular}{llcccccccc}
\toprule
crit. & ckpt & minimiser & rounds & q.\ imp. & q.\ stop & updates & P(found) & speedup (unit) & speedup (hw) \\
\midrule
A & 200 & L25b\_30 & 77.5 & 7.5 & 228.9 & 3.10 & 1.00 & 0.21$\times$ & 1.04$\times$ \\
A & 400 & corner\_low & 78.2 & 7.2 & 227.8 & 3.30 & 1.00 & 0.21$\times$ & 1.04$\times$ \\
A & 600 & corner\_low & 78.5 & 7.3 & 226.2 & 3.45 & 1.00 & 0.21$\times$ & 1.03$\times$ \\
A & 800 & corner\_low & 78.8 & 7.7 & 226.6 & 3.45 & 1.00 & 0.21$\times$ & 1.03$\times$ \\
A & 1000 & corner\_low & 77.1 & 6.2 & 230.8 & 2.95 & 1.00 & 0.21$\times$ & 1.05$\times$ \\
B & 200 & L25b\_30 & 27.5 & 7.5 & 29.4 & 3.10 & 1.00 & 0.88$\times$ & 2.95$\times$ \\
B & 400 & corner\_low & 27.4 & 7.2 & 29.0 & 3.30 & 1.00 & 0.89$\times$ & 2.96$\times$ \\
B & 600 & corner\_low & 27.8 & 7.3 & 29.1 & 3.45 & 1.00 & 0.88$\times$ & 2.91$\times$ \\
B & 800 & corner\_low & 28.5 & 7.7 & 28.9 & 3.45 & 1.00 & 0.87$\times$ & 2.84$\times$ \\
B & 1000 & corner\_low & 25.7 & 6.2 & 29.1 & 2.95 & 1.00 & 0.93$\times$ & 3.15$\times$ \\
C & 200 & L25b\_30 & 50.7 & 7.5 & 121.3 & 3.10 & 1.00 & 0.35$\times$ & 1.60$\times$ \\
C & 400 & corner\_low & 50.8 & 7.2 & 121.2 & 3.30 & 1.00 & 0.35$\times$ & 1.59$\times$ \\
C & 600 & corner\_low & 51.2 & 7.3 & 121.3 & 3.45 & 1.00 & 0.35$\times$ & 1.58$\times$ \\
C & 800 & corner\_low & 51.8 & 7.7 & 120.7 & 3.45 & 1.00 & 0.35$\times$ & 1.56$\times$ \\
C & 1000 & corner\_low & 48.9 & 6.2 & 121.3 & 2.95 & 1.00 & 0.36$\times$ & 1.66$\times$ \\
D & 200 & L25b\_30 & 26.8 & 7.5 & 26.6 & 3.10 & 1.00 & 0.92$\times$ & 3.02$\times$ \\
D & 400 & corner\_low & 26.8 & 7.2 & 25.8 & 3.30 & 1.00 & 0.94$\times$ & 3.02$\times$ \\
D & 600 & corner\_low & 27.1 & 7.3 & 25.2 & 3.45 & 1.00 & 0.94$\times$ & 2.99$\times$ \\
D & 800 & corner\_low & 27.8 & 7.7 & 25.5 & 3.45 & 1.00 & 0.91$\times$ & 2.92$\times$ \\
D & 1000 & corner\_low & 25.1 & 6.2 & 25.9 & 2.95 & 1.00 & 0.99$\times$ & 3.23$\times$ \\
\bottomrule
\end{tabular}
}
\end{center}
Figure~\ref{fig:selection-time} prices the same counts per checkpoint under the two conventions of
Appendix~\ref{app:cost}. Under the unit-cost query model no criterion reaches parity at this size:
$Q^{(x)}+2R^{(x)}$ is at least $90$ against $N=81$, giving $S_{\mathrm{unit}}$ of $0.21$, $0.84$, $0.35$ and
$0.90$ for A to D. The dotted line marks the bound of Eq.~\eqref{eq:dhbound} in the same units.
Under the hardware-informed estimate every bar collapses onto its checking-cost floor $R^{(x)}\,t_{\mathrm{chk}}$, and $S_{\mathrm{hw}}$ is $1.05$, $2.77$, $1.55$ and $2.86$; the quantum
terms are invisible at this scale, as follows from Eq.~\eqref{eq:shw}. Both readings are conditional on
the oracle convention of Appendix~\ref{app:cost}. Criterion D carries the scaling study.

\begin{figure}[!htb]
  \centering
  \includegraphics[width=\textwidth]{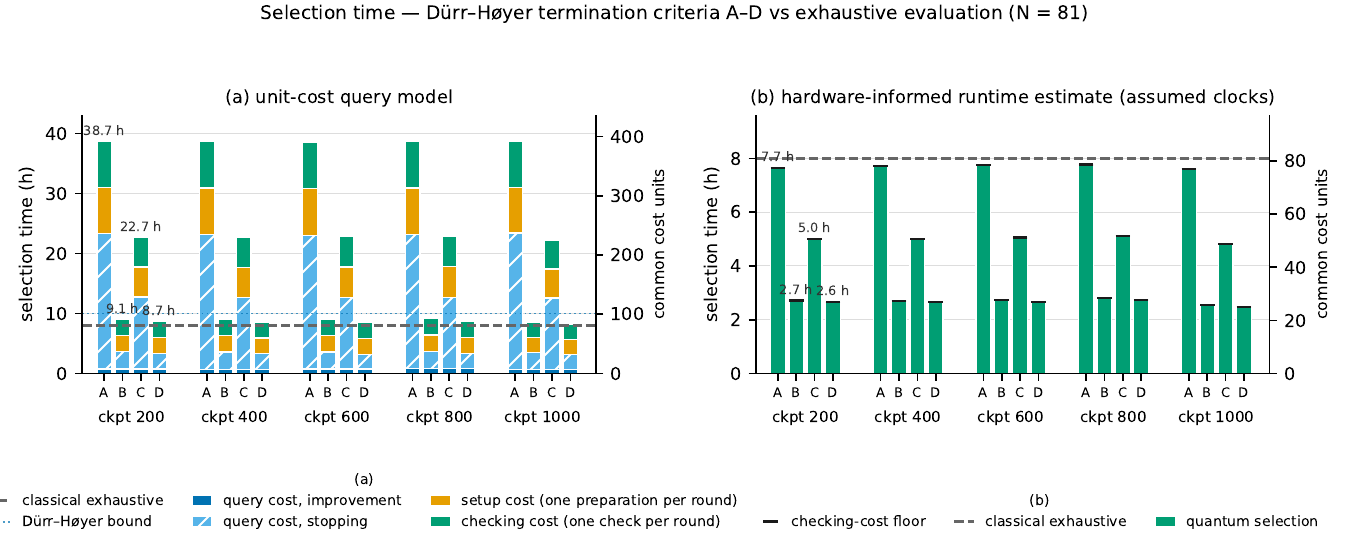}
  \caption{Selection time per checkpoint for the four termination criteria on the $h=25$ tables,
  $N=81$, against exhaustive evaluation (dashed, $81\times355.5$~s $=8.00$~h). One round is one
  state preparation, $j$ Grover iterations, one measurement and one classical check, so
  preparations and checks both appear as the round count $R^{(x)}$. (a) Unit-cost query model, $t_{\mathrm{iter}}=t_{\mathrm{setup}}=t_{\mathrm{chk}}$: bars stack the improvement and stopping
  iterations, the preparations and the checks; the right axis reads the bar in common cost units,
  and the dotted line is $\tfrac{45}{4}\sqrt N\approx101$ units. (b) Hardware-informed estimate, drawn with
  the interim clocks $t_{\mathrm{iter}}=1$~ms and $t_{\mathrm{setup}}=40~\mu$s and the checks at
  $t_{\mathrm{chk}}$; the clocks of Eq.~\eqref{eq:hwclock} are smaller and change no bar: every bar
  equals its checking-cost floor $R^{(x)}\,t_{\mathrm{chk}}$. All timings are
  conditional on the lookup-oracle convention; panel (b) is never a measured speedup.}
  \label{fig:selection-time}
\end{figure}

\clearpage
\section{Declared-rule continuation and its agreement with the measured rows}
\label{app:scaling}
The declared-rule cross-check of Section~\ref{sec:setup} replays criterion D with the analytic
probability of Eq.~\eqref{eq:grover}, no table and no circuit. The first table repeats the measured rows at full width; the second continues the rule beyond
the measured range; every row of it is a cross-check of the rule and not a measurement, and the
times assume $t_{\mathrm{chk}}=355.5$~s and the interim clocks $1$~ms, $40~\mu$s (Eq.~\ref{eq:hwclock} lowers both and moves no entry). The second table is the
validation gate at the three measured sizes: the pass band is $|\text{Aer}-\text{model}|\le
\max(3\,\text{s.e.},\,3\%)$ on both $R^{(\mathrm D)}$ and $Q^{(\mathrm D)}$, and all three sizes pass.
{\scriptsize\par\noindent\textit{Part A, measured rows (full width)}\par\smallskip
\resizebox{\textwidth}{!}{\begin{tabular}{llccccccccccc}
\toprule
table size $N$ & rows & trials (Aer runs or model runs) & rounds (prep.\ = checks) $\pm$ s.e. & queries (total) $\pm$ s.e. & thr.\ updates (diag.) & P(minimizer found) & classical exhaustive & quantum, unit-cost & speedup (unit-cost) & quantum, hardware-informed & speedup (hardware-informed) \\
\midrule
81 & measured (Aer) & 5,000 Aer runs (5 $\times$ 1000) & 28.3 $\pm$ 0.1 & 33.8 $\pm$ 0.1 & 3.95 & 0.999 & 8.0 h & 8.9 h & 0.90$\times$ & 2.8 h & 2.86$\times$ \\
128 & measured (Aer) & 5,000 Aer runs (5 $\times$ 1000) & 33.3 $\pm$ 0.1 & 46.6 $\pm$ 0.1 & 4.40 & 1.000 & 12.6 h & 11.2 h & 1.13$\times$ & 3.3 h & 3.85$\times$ \\
256 & measured (Aer) & 5,000 Aer runs (1 $\times$ 5000) & 41.5 $\pm$ 0.2 & 71.8 $\pm$ 0.2 & 5.10 & 1.000 & 25.3 h & 15.3 h & 1.65$\times$ & 4.1 h & 6.17$\times$ \\
\bottomrule
\end{tabular}}\par\medskip

\par\noindent\textit{Part B, declared-rule continuation (not a measurement)}\par\smallskip
\resizebox{\textwidth}{!}{\begin{tabular}{llccccccccccc}
\toprule
table size $N$ & rows & model runs & rounds $\pm$ s.e. & queries (total) $\pm$ s.e. & thr.\ updates (diag.) & P(minimizer found) & classical exhaustive & quantum, unit-cost & speedup (unit-cost) & quantum, hardware-informed & speedup (hardware-informed) \\
\midrule
512 & theory cross-check (declared rule) & 2,000 model runs (no table, no circuits) & 51.0 $\pm$ 0.3 & 110.3 $\pm$ 0.6 & 5.84 & 1.000 & 2.1 d & 21.0 h & 2.41$\times$ & 5.0 h & 10.0$\times$ \\
1,024 & theory cross-check (declared rule) & 2,000 model runs (no table, no circuits) & 61.4 $\pm$ 0.4 & 164.0 $\pm$ 0.8 & 6.56 & 1.000 & 4.2 d & 28.3 h & 3.57$\times$ & 6.1 h & 16.7$\times$ \\
4,096 & theory cross-check (declared rule) & 2,000 model runs (no table, no circuits) & 86.4 $\pm$ 0.6 & 362.4 $\pm$ 1.9 & 7.89 & 1.000 & 16.9 d & 2.2 d & 7.65$\times$ & 8.5 h & 47.4$\times$ \\
16,384 & theory cross-check (declared rule) & 2,000 model runs (no table, no circuits) & 117.1 $\pm$ 0.7 & 770.5 $\pm$ 4.0 & 9.26 & 1.000 & 67.4 d & 4.1 d & 16.3$\times$ & 11.6 h & 140$\times$ \\
65,536 & theory cross-check (declared rule) & 1,000 model runs (no table, no circuits) & 154.5 $\pm$ 1.3 & 1599.1 $\pm$ 11.9 & 10.69 & 1.000 & 269.7 d & 7.9 d & 34.3$\times$ & 15.3 h & 424$\times$ \\
262,144 & theory cross-check (declared rule) & 1,000 model runs (no table, no circuits) & 196.9 $\pm$ 1.6 & 3253.3 $\pm$ 23.2 & 12.24 & 1.000 & 2.96 yr & 15.0 d & 71.9$\times$ & 19.4 h & 1,331$\times$ \\
\bottomrule
\end{tabular}}\par\medskip

\par\noindent\textit{Part C, agreement at the measured sizes}\par\smallskip
\resizebox{\textwidth}{!}{\begin{tabular}{lccccc}
\toprule
table size $N$ & rounds, Aer & rounds, model & queries, Aer & queries, model & thr.\ updates Aer / model (diag.) \\
\midrule
81 & 28.3 & 28.4 $\pm$ 0.1 & 33.8 & 33.7 $\pm$ 0.1 & 3.95 / 3.95 \\
128 & 33.3 & 33.2 $\pm$ 0.1 & 46.6 & 46.5 $\pm$ 0.2 & 4.40 / 4.43 \\
256 & 41.5 & 41.3 $\pm$ 0.2 & 71.8 & 71.5 $\pm$ 0.2 & 5.10 / 5.12 \\
\bottomrule
\end{tabular}}\par\medskip
}
\begin{table}[!htb]
  \centering
  \caption{Selection cost against table size for criterion D ($\delta=10^{-3}$), measured rows only:
  ideal Grover circuits simulated on Qiskit Aer over the real $h=25$ scores. Trials are checkpoints
  times trials per checkpoint; rounds and queries carry standard errors; the speedups follow
  Eqs.~\eqref{eq:sunit} and \eqref{eq:shw} with $t_{\mathrm{chk}}=355.5$~s and the interim clocks $1$~ms, $40~\mu$s (Eq.~\eqref{eq:hwclock} lowers both and moves no entry). The full-width rows (threshold updates, classical and quantum times),
  the declared-rule continuation beyond $N=256$ and its agreement with these rows are
  given in this appendix.}
  \label{tab:scaling}
  \footnotesize
  \resizebox{\textwidth}{!}{\begin{tabular}{lcccccc}
\toprule
size & trials & rounds & queries & P(min.\ found) & speedup (unit-cost) & speedup (hw-informed) \\
\midrule
81 & 5 $\times$ 1,000 & 28.3 $\pm$ 0.1 & 33.8 $\pm$ 0.1 & 0.999 & 0.90$\times$ & 2.86$\times$ \\
128 & 5 $\times$ 1,000 & 33.3 $\pm$ 0.1 & 46.6 $\pm$ 0.1 & 1.000 & 1.13$\times$ & 3.85$\times$ \\
256 & 1 $\times$ 5,000 & 41.5 $\pm$ 0.2 & 71.8 $\pm$ 0.2 & 1.000 & 1.65$\times$ & 6.17$\times$ \\
\bottomrule
\end{tabular}
}
\end{table}

\begin{figure}[!htb]
  \centering
  \includegraphics[width=0.8\textwidth]{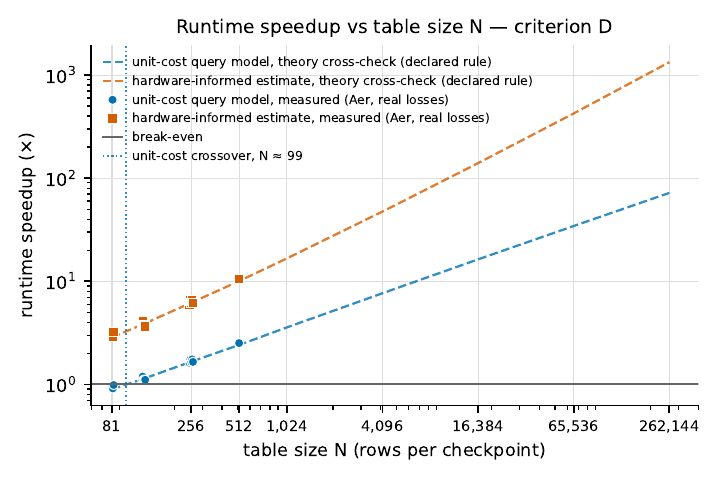}
  \caption{Runtime speedup versus table size $N$ for criterion D with $\delta=10^{-3}$. Dashed curves are the
  declared-rule cross-check obtained from the analytic success probability of Eq.~\eqref{eq:grover}, and are
  continued to the design size $N=262{,}144$, where no corresponding table or circuit experiment is run. At
  the measured sizes, the cross-check agrees with the measured results within $0.5\%$. The construction of the
  score table is not included.}
  \label{fig:speedup-N}
\end{figure}

Criterion D is followed across the three measured sizes in Figure~\ref{fig:speedup-N}. At unit cost, one
per iteration and two per round, the speedup over exhaustive evaluation crosses one between $N=81$ and
$128$: $S_{\mathrm{unit}}$ is $(0.90,\,1.13,\,1.65)$ and the break-even sits near $N\approx100$. Under
the hardware-informed estimate the classical checks dominate and $S_{\mathrm{hw}}$ is
$(2.86,\,3.85,\,6.14)$, the floor $N/R$ at every size: the quantum terms are negligible, one selection
costs its $R$ classical checks, and the gain comes from needing fewer rounds rather than from a faster
circuit.
Both ratios exclude the
construction of $\Tscore$: with each table charged at one replay per profile, a selection over the lookup
oracle takes longer than exhaustive evaluation at every measured size, the ratio of exhaustive to quantum
time being $0.74$, $0.79$ and $0.86$ at $N=81$, $128$ and $256$, so the lookup route carries no end-to-end
advantage.

The declared-rule cross-check of Section~\ref{sec:setup} reproduces the measured rows within
$0.5\%$ in both $R^{(\mathrm D)}$ and $Q^{(\mathrm D)}$ at all three sizes, which is the license
to draw the same rule beyond the measured range as the dashed curves of Figure~\ref{fig:speedup-N}.
Continued to the design size $N=8^6=262{,}144$ the rule gives $R^{(\mathrm D)}\approx197$ rounds and $Q^{(\mathrm D)}\approx3{,}250$ iterations, an $S_{\mathrm{unit}}$ of about $72$ and an $S_{\mathrm{hw}}$ of about
$1{,}330$. Along the ladder $Q^{(\mathrm D)}$ grows as $\sqrt N$ ($Q^{(\mathrm D)}/\sqrt N$ from $4.1$ to $6.4$, inside the $11.25$ of Eq.~\ref{eq:dhbound}) and $R^{(\mathrm D)}$ as $(\ln N)^2$, so the floor $N/R^{(\mathrm D)}$ grows almost linearly
in $N$. Nothing beyond $N=256$ is a measurement: the continuation assumes distinct scores and a
uniform rank of the initial threshold, and, like every selection-time estimate of the lookup study, it is
conditional on the lookup-oracle convention.

\clearpage
\section{Seed-stability check at checkpoint 200}
\label{app:seeds}
Criterion D at checkpoint 200 for $N=81$, $128$ and $256$, with the paper's seed block (block 0)
and four further disjoint blocks of $1{,}000$ trials. The spread of the five block means is
$0.62/0.94/1.19$ rounds and $0.63/1.02/1.50$ queries at the three sizes, i.e.\ $2.6$ to $3.7$
standard errors, the range expected for five independent blocks; no block stands out. At $N=256$
block 0 is the $5{,}000$-trial file and therefore contains blocks 1--4; blocks 1--4 alone carry the
independent statement.
\begin{center}
  \small
  \resizebox{\textwidth}{!}{\begin{tabular}{rrlrrrrr}
\toprule
$N$ & block & trial seeds & trials & $R\pm$ s.e. & $Q\pm$ s.e. & P(minimiser found) & misses \\
\midrule
81 & 0 & 1000-1999 & 1,000 & 28.61 $\pm$ 0.25 & 33.98 $\pm$ 0.24 & 0.999 & 1 \\
81 & 1 & 2000-2999 & 1,000 & 27.99 $\pm$ 0.23 & 33.35 $\pm$ 0.23 & 1.000 & 0 \\
81 & 2 & 3000-3999 & 1,000 & 28.11 $\pm$ 0.23 & 33.46 $\pm$ 0.23 & 1.000 & 0 \\
81 & 3 & 4000-4999 & 1,000 & 28.36 $\pm$ 0.24 & 33.63 $\pm$ 0.24 & 1.000 & 0 \\
81 & 4 & 5000-5999 & 1,000 & 28.51 $\pm$ 0.25 & 33.65 $\pm$ 0.24 & 1.000 & 0 \\
128 & 0 & 1000-1999 & 1,000 & 33.72 $\pm$ 0.29 & 47.03 $\pm$ 0.35 & 1.000 & 0 \\
128 & 1 & 2000-2999 & 1,000 & 33.42 $\pm$ 0.29 & 46.77 $\pm$ 0.33 & 0.999 & 1 \\
128 & 2 & 3000-3999 & 1,000 & 32.97 $\pm$ 0.28 & 46.01 $\pm$ 0.32 & 0.998 & 2 \\
128 & 3 & 4000-4999 & 1,000 & 33.29 $\pm$ 0.30 & 46.40 $\pm$ 0.33 & 1.000 & 0 \\
128 & 4 & 5000-5999 & 1,000 & 32.78 $\pm$ 0.29 & 46.22 $\pm$ 0.35 & 0.998 & 2 \\
256 & 0 & 1000-5999 & 5,000 & 41.52 $\pm$ 0.16 & 71.78 $\pm$ 0.22 & 1.000 & 2 \\
256 & 1 & 2000-2999 & 1,000 & 41.93 $\pm$ 0.36 & 72.48 $\pm$ 0.51 & 0.999 & 1 \\
256 & 2 & 3000-3999 & 1,000 & 42.04 $\pm$ 0.37 & 72.21 $\pm$ 0.50 & 1.000 & 0 \\
256 & 3 & 4000-4999 & 1,000 & 40.85 $\pm$ 0.35 & 70.98 $\pm$ 0.48 & 1.000 & 0 \\
256 & 4 & 5000-5999 & 1,000 & 41.31 $\pm$ 0.36 & 71.75 $\pm$ 0.50 & 1.000 & 0 \\
\bottomrule
\end{tabular}
}
\end{center}

\clearpage
\section{Interaction-Guided Schedule Oracle: score construction and closed-loop runs}
\label{app:igso}
\paragraph{Checkpoint evaluation.} At each checkpoint $t$, the checkpoint evaluation of Section~\ref{sec:igso} measures change
patterns around the uniform-rate baseline $z_{\mathrm{ref}}$. A change pattern $\gamma$ moves one, two or three groups away from
$z_{\mathrm{ref}}$ and keeps the others at $1.0$. For each pattern, the parameters are moved once from the saved model and
optimizer state along the optimizer's update direction for one fixed batch of $16$ training sequences, the update of each group
scaled by its rate, and $\Lambda_t(\gamma)$ is the loss of the moved model on two fixed held-out sequences; the baseline takes
the same step from the same state. The gain of $\gamma$ is $Y_t(\gamma)=\Lambda_t(z_{\mathrm{ref}})-\Lambda_t(\gamma)$, and
$Y_t(g;\eta)$ is the gain of moving group $g$ alone to the rate $\eta$. Every term of Eq.~\eqref{eq:gainmodel} vanishes when one
of its groups stays at $1.0$, so the baseline scores zero.

\paragraph{Interaction coefficients.} For a change pattern $\gamma$, the interaction term is the
discrete derivative of the gain over the moved groups \citep{sundararajan2020shapleytaylor},
\begin{equation}
  \nu_t(\gamma)=\sum_{\gamma'\subseteq\gamma}(-1)^{|\gamma|-|\gamma'|}\,Y_t(\gamma'),
  \label{eq:dividend}
\end{equation}
where $\gamma'$ runs over the sub-patterns of $\gamma$, each moved group keeping its rate in $\gamma$, the empty one with gain
zero. For one group, a pair and a triple, with the rates omitted,
\begin{align*}
  \nu_{t,g}&=Y_{t,g},\qquad \nu_{t,gg'}=Y_{t,gg'}-Y_{t,g}-Y_{t,g'},\\
  \nu_{t,gg'g''}&=Y_{t,gg'g''}-Y_{t,gg'}-Y_{t,gg''}-Y_{t,g'g''}+Y_{t,g}+Y_{t,g'}+Y_{t,g''},
\end{align*}
where $Y_{t,gg'}$ is the measured gain of moving $g$ and $g'$ together: a pair or triple interaction is the part of a measured
joint gain that its lower-order terms do not explain, derived after the joint measurement rather than measured on its own.
Summed over the sub-patterns of a measured pattern, the terms of Eq.~\eqref{eq:gainmodel} therefore return its measured gain
before rounding. For a schedule with $0.7$ in a pair or triple, or with more than three moved groups, $\widehat Y_t(z)$ is a
modeled gain rather than a measured joint gain. Taken at the uniform-rate baseline only, these terms are not Shapley--Taylor
indices, whose highest-order indices average over orderings.

\paragraph{Design and coefficients.} At each checkpoint the measurement design has $249$ rows: five repeats of the
uniform-rate baseline, the $12$ single-group changes to $0.5$ and $1.4$ times the uniform rate, $12$ single-group
changes to $0.7$ and $2.0$ times that rate, the $60$ pairs and the $160$ triples of Section~\ref{sec:igso} at $0.5$ and $1.4$.
The rows at $2.0$ enter no coefficient, and the search menu does not contain that rate. Over the moved rates $0.5$, $0.7$ and
$1.4$, the three sums of Eq.~\eqref{eq:gainmodel} can draw on $6\times3=18$ single, $\binom62\times3^2=135$ pair and
$\binom63\times3^3=540$ triple cases, $693$ in all; these count possible rate-specific cases, not stored coefficients, of which
the design holds the $232$ single, pair and triple patterns at $0.5$ and $1.4$ and at most six corrections at $0.7$.
A group at $0.7$ keeps its own measured single gain, through the correction of Eq.~\eqref{eq:igscore}, and shares the pair and
triple interactions of its lowering to $0.5$. The $15\times5=75$ pair and $20\times19=380$ triple cases that contain $0.7$ have
no measured row: sharing the interactions at $0.5$ avoids these $455$ joint measurements per checkpoint, which would raise the
design from $249$ to $704$ rows, and lets each pair or triple coefficient act on the direction of its groups, lowered or
raised, rather than on separate rate-specific terms. The unit $\varepsilon$ is the larger
of the spread of the uniform repeats and the declared $2\times10^{-4}$; the repeats are identical within one
measurement, so $\varepsilon=2\times10^{-4}$ in both runs. The five-batch run measures the same rows on five fixed
batches of $16$ training sequences, the first being the single batch, averages the losses row by row and records
the spread of each row without using it. The single-batch measurements were taken with unseeded dropout and
non-deterministic GPU kernels and do not repeat exactly: a repeat of the same measurement on one checkpoint
changes the losses by $6.8\times10^{-4}$ on average; the five-batch run used seeded dropout and deterministic
kernels.

\paragraph{Integer score.} Each coefficient is divided by $\varepsilon$ before it is rounded, which enlarges the small gains
before the rounding: $a_t(\gamma)=\mathrm{round}\big(\nu_t(\gamma)/\varepsilon\big)$ for each of the $232$ patterns, and the
$0.7$ correction of group $g$ is $b_t(g)=\mathrm{round}\big((Y_t(g;0.7)-Y_t(g;0.5))/\varepsilon\big)$, so that a group at
$0.7$ carries its own measured single gain. With $z\supseteq\gamma$ when $z$ moves every group of $\gamma$ in the same
direction as $\gamma$ ($0.5$ and $0.7$ counting as lowered) and $\Gamma$ the set of the $232$ patterns, the score evaluated by
the circuit and by the classical check of each sample is
\begin{equation}
  \Phi_t(z)=\sum_{\gamma\in\Gamma}a_t(\gamma)\,\mathbf 1[z\supseteq\gamma]+\sum_{g=1}^{G}b_t(g)\,\mathbf 1[\eta_g=0.7].
  \label{eq:igscore}
\end{equation}
At checkpoint $200$ of the single-batch run, for example, the second layer group has the single gain $3.69\times10^{-4}$ at
$0.5$, which rounds to zero directly but to $a_t=2$ after division by $\varepsilon$ ($1.85$), and the gain $1.14\times10^{-3}$
at $0.7$ gives $b_t=4$, so the group contributes $6$ at $0.7$. Patterns whose rounded coefficient is zero add nothing and are
left out of the circuit. Because the coefficients and corrections are rounded separately, $\Phi_t(z)$ need not equal
$\widehat Y_t(z)/\varepsilon$ rounded once.

\paragraph{Around another baseline.} Eq.~\eqref{eq:genbase} of Section~\ref{sec:summary} writes the score around a baseline
schedule $z_{\mathrm{base}}$ as single-group gains plus pair and triple interactions. With $z_{\mathrm{base}}\oplus\gamma$ the
baseline with only the groups of a pattern $\gamma$ moved to the rates of $z$, and $\mathcal P(z\mid z_{\mathrm{base}})$ the
patterns of one to three groups on which $z$ differs from $z_{\mathrm{base}}$, the inclusion--exclusion of
Eq.~\eqref{eq:dividend} around the baseline gives
\begin{equation}
  \widehat Y_t(z\mid z_{\mathrm{base}})=\sum_{\gamma\in\mathcal P(z\mid z_{\mathrm{base}})}\ \sum_{\gamma'\subseteq\gamma}
  (-1)^{|\gamma|-|\gamma'|}\big[\Lambda_t(z_{\mathrm{base}})-\Lambda_t(z_{\mathrm{base}}\oplus\gamma')\big].
  \label{eq:genbase-ie}
\end{equation}
For a pattern of one, two or three groups the inner sum is the single-group gain, the pair interaction or the triple
interaction around $z_{\mathrm{base}}$, so Eq.~\eqref{eq:genbase-ie} is Eq.~\eqref{eq:genbase} term by term: every group outside a
pattern keeps its baseline rate, every term vanishes when one of its groups stays at its baseline rate, and the
baseline scores zero. The terms are rate-specific effects measured around $z_{\mathrm{base}}$; the sharing of the $0.5$
interactions at $0.7$ is a choice of the reported runs and is not part of this expression.

\begin{figure}[!htb]
  \centering
  \includegraphics{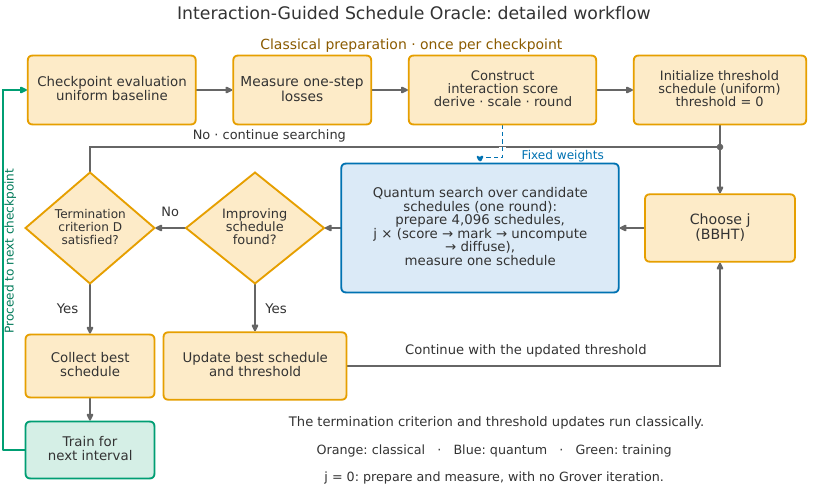}
  \caption{Detailed workflow of the Interaction-Guided Schedule Oracle at one checkpoint. Classical loss measurements define
  fixed integer coefficients, and the best schedule is initialized at the uniform-rate baseline. Each quantum round prepares
  $4{,}096$ schedules, applies the chosen number of oracle and diffusion iterations, and measures one schedule. Classical
  checking updates the best schedule and threshold after an improvement; otherwise criterion D determines whether to continue.
  After stopping, the best schedule controls the next training interval, and the next checkpoint rebuilds the evidence.}
  \label{fig:igso-detailed}
\end{figure}

\paragraph{Circuit and search.} The register holds two qubits per group and an accumulator of
$\lceil\log_2(\text{span}+1)\rceil+2$ qubits, where the span is the range of $\Phi_t$, so $22$ to $25$ qubits in
all. The accumulator, in the Fourier basis, receives each $a_t(\gamma)$ and $b_t(g)$ as a phase rotation controlled on the
qubits of the groups involved; subtracting $\Phi_t(y)+1$, flipping the phase of the sign qubit and uncomputing the arithmetic
realize Eq.~\eqref{eq:igmark} on every schedule at once, and the circuit returns no score as a readable value. The search is
the minimum finding of \citet{durr1996} with a larger score being better. Each round prepares the joint schedules in
superposition, applies the oracle and diffusion the number of times drawn by Eq.~\eqref{eq:bbht}, and samples one schedule; a
classical evaluation of Eq.~\eqref{eq:igscore} on the sample updates the best schedule and the threshold after an improvement,
and after an unsuccessful sample criterion D, the criterion chosen in Section~\ref{sec:res-criteria}, decides whether to
continue. The uniform-rate baseline is returned if nothing beats it. Before each run, the circuit was checked on the checkpoint-$200$
coefficients: at the initial threshold its marks equal those of Eq.~\eqref{eq:igmark} evaluated with Eq.~\eqref{eq:igscore} on
all $4{,}096$ schedules, and the accumulator returns to zero.
This enumeration exists only in the test; the search never reads it. Each search is one trial with its random
seed set to the checkpoint step. Figure~\ref{fig:igso-detailed} shows the workflow of one checkpoint at this level of detail.
Criterion D is reused here as a stopping rule. Its rank distribution (Appendix~\ref{app:critA}) assumes a first
threshold drawn uniformly at random and distinct scores, whereas this search starts from the fixed uniform-rate
baseline and the integer scores of Eq.~\eqref{eq:igscore} can tie; the rule was applied unchanged, so
$\delta=10^{-3}$ is a nominal setting and not a calibrated error level, and the returned schedule is not
certified as the maximizer of $\Phi_t$. The single-batch searches ran on a
laptop CPU in single precision ($4{,}465$ to $16{,}362$~s of simulator time per search), the five-batch searches
on a GPU in double precision ($158$ to $281$~s); these are times of the classical simulator, not of a quantum
device.

\paragraph{Decisions.} Table~\ref{tab:igso-decisions} lists, for each leg, the schedule returned (group order from
the first layers to the last), its interaction score $\Phi$ in units of $\varepsilon$, the rounds $R$ and Grover
iterations $Q$ of the search, the register width, and, for the five-batch run, the number of the $232$
interaction terms whose sign differs between the first batch and the five-batch average. Both runs start from the checkpoint at
step $200$ of the uniform-rate run of seed $2$; each interval is $200$ steps ($221$ for the last). Besides the evidence, the
two runs also differ in the GPU, the simulator precision and the determinism of the training kernels.
\begin{table}[!htb]
  \centering
  \caption{The five decisions of each closed-loop run. The scores $\Phi$ of the two runs are not comparable
  with each other, since the averaged coefficients are smaller.}
  \label{tab:igso-decisions}
  \footnotesize
  \resizebox{\textwidth}{!}{\begin{tabular}{l lrrrr lrrrrr}
\toprule
 & \multicolumn{5}{c}{single-batch measurements} & \multicolumn{6}{c}{five-batch measurements} \\
\cmidrule(lr){2-6}\cmidrule(lr){7-12}
leg & schedule & $\Phi$ & $R$ & $Q$ & qubits & schedule & $\Phi$ & $R$ & $Q$ & qubits & sign flips \\
\midrule
$200\to400$ & (0.5, 1.4, 1.4, 0.5, 0.7, 0.5) & 129 & 87 & 264 & 24 & (0.5, 1.0, 1.4, 0.7, 1.4, 0.7) & 12 & 62 & 271 & 22 & 106 \\
$400\to600$ & (0.5, 1.4, 0.7, 0.7, 1.4, 1.4) & 66 & 102 & 296 & 24 & (1.4, 0.7, 0.7, 0.7, 0.5, 1.4) & 25 & 75 & 320 & 22 & 76 \\
$600\to800$ & (1.4, 0.5, 1.4, 0.7, 1.4, 1.4) & 78 & 109 & 494 & 24 & (0.7, 0.5, 0.5, 0.5, 1.4, 0.5) & 33 & 77 & 289 & 22 & 79 \\
$800\to1000$ & (1.4, 1.4, 0.5, 0.7, 1.4, 1.4) & 162 & 61 & 292 & 25 & (1.4, 0.7, 1.4, 0.5, 1.4, 0.5) & 11 & 52 & 269 & 23 & 73 \\
$1000\to1221$ & (0.7, 1.4, 0.5, 1.4, 1.0, 1.4) & 28 & 76 & 304 & 24 & (1.4, 1.4, 0.5, 1.4, 0.5, 0.7) & 32 & 76 & 292 & 22 & 88 \\
\bottomrule
\end{tabular}
}
\end{table}

\ifibmloop
\clearpage
\section{Hardware-constrained Grover iteration counts}
\label{app:ibm-jlimit}
\paragraph{Compiled gate accounting.}
For the hardware-sampled searches of Section~\ref{sec:res-ibmloop}, the circuit of a round with $j$ Grover iterations is
assembled from compiled pieces: a prefix that prepares the uniform superposition on the $q_{\mathrm{srch}}=12$ search qubits,
then $j$ iteration bodies, then the measurement of the search register. Body $j'$ is one Grover iteration, the oracle of
Eq.~\eqref{eq:igmark} followed by the diffuser, routed on the device starting from the qubit placement that body $j'-1$ left,
so the bodies are not identical and differ in gate count; the submitted circuit is the concatenation of the cached
compiled pieces, with no further compilation. For each gate type,
\begin{equation}
  C_{\mathrm{CZ}}(j)=C_{\mathrm{CZ}}(0)+\sum_{j'=1}^{j}\Delta C_{\mathrm{CZ}}(j'),
  \label{eq:gatecount}
\end{equation}
and likewise for RZ and SX, the components of $\mathbf C$ in Eq.~\eqref{eq:jdev}. Each Hadamard gate of the prefix is compiled
as RZ, SX, RZ, so $\mathbf C(0)=(0,\,24,\,12)$; the measurement adds no gate of the three types. The counts depend on the
checkpoint, through the integer coefficients of Eq.~\eqref{eq:igscore}, on the threshold, and on the backend, synthesis,
layout and routing used at compile time, and they are recomputed for each. On assembled circuits at checkpoint $200$, with
the threshold score $\Phi_t(y)=42$ on \texttt{ibm\_fez}, the gate counts equal the sums of Eq.~\eqref{eq:gatecount} exactly:
$1{,}799{,}810$ CZ, $2{,}348{,}632$ RZ and $3{,}520{,}886$ SX gates at $j=28$, and $3{,}478{,}853$, $4{,}525{,}511$ and
$6{,}797{,}218$ at $j=54$.

\paragraph{The limit.}
IBM specifies per-circuit limits of $5\times10^{6}$ two-qubit gates, $3\times10^{7}$ RZ gates and $2\times10^{7}$ SX gates
(IBM Quantum documentation, ``Job limits'', \url{https://quantum.cloud.ibm.com/docs/en/guides/job-limits}, read
11~November~2026). With $\mathbf C(0)$ the numbers of CZ, RZ and SX gates of the preparation and $\Delta\mathbf C(j')$
those added by the $j'$-th body, the largest iteration count within the per-circuit gate limits of the service is
\begin{equation}
  J_{\mathrm{dev}}=\max\Big\{\,j\in\mathbb Z_{\ge0}:\ \mathbf C(0)+\sum_{j'=1}^{j}\Delta\mathbf C(j')\le(5,\,30,\,20)\times10^{6}\Big\},
  \label{eq:jdev}
\end{equation}
the inequality holding for each gate type, recomputed for every checkpoint, threshold and backend. The compiled circuits are
CZ-native, so Eq.~\eqref{eq:jdev} applies these limits to the counts of
Eq.~\eqref{eq:gatecount}. The maximum in Eq.~\eqref{eq:jdev} runs over all non-negative integers, so $J_{\mathrm{dev}}$ is a
property of the compiled circuit family and the device limits and does not depend on the range of the BBHT draw. Every body
adds gates, so the counts never decrease in $j$, the admissible counts form the interval $\{0,\dots,J_{\mathrm{dev}}\}$, and
$J_{\mathrm{dev}}$ is found by compiling bodies until the first one crosses a limit. Two evaluations exist in the program and
are kept apart. A measurement stage compiles until a limit is crossed and returns $J_{\mathrm{dev}}$ exactly. During the
search, bodies are compiled only as far as the current block of rounds can draw, at most $\lceil\sqrt N\rceil-1=63$ at
$N=4{,}096$; if all of them fit, the search records only $J_{\mathrm{dev}}\ge63$ and the bound is inactive. The largest count
that the schedule can draw, $63$, is a property of the schedule and a different number from $J_{\mathrm{dev}}$.

\paragraph{Estimate before compilation.}
Before the whole range is compiled, a first-order projection follows from the fixed counts and a representative
per-iteration count $\overline{\Delta C}$ of each gate type,
\begin{equation}
  \widehat J_{\mathrm{dev}}=\min\bigg\{\bigg\lfloor\frac{5\times10^{6}-C_{\mathrm{CZ}}(0)}{\overline{\Delta C}_{\mathrm{CZ}}}\bigg\rfloor,\
  \bigg\lfloor\frac{3\times10^{7}-C_{\mathrm{RZ}}(0)}{\overline{\Delta C}_{\mathrm{RZ}}}\bigg\rfloor,\
  \bigg\lfloor\frac{2\times10^{7}-C_{\mathrm{SX}}(0)}{\overline{\Delta C}_{\mathrm{SX}}}\bigg\rfloor\bigg\}.
  \label{eq:jdev-est}
\end{equation}
It is a planning estimate only; every decision of the search uses the exact cumulative counts of Eqs.~\eqref{eq:gatecount}
and~\eqref{eq:jdev}. Table~\ref{tab:jdev} lists the compilations of the same oracle at checkpoint $200$ with the threshold
score $42$ on \texttt{ibm\_fez}. In both measured cases the estimate equals the exact value, and CZ is the binding gate type.
For the shared-AND synthesis $J_{\mathrm{dev}}=77$, above the schedule maximum of $63$, so every circuit the schedule can draw at
this threshold is inside the three counted limits, the largest, $j=63$, holding $4{,}062{,}940$ CZ, $5{,}287{,}620$ RZ and
$7{,}940{,}324$ SX gates. The shared-AND synthesis is the same phase oracle built from shared AND indicators and
compute--diagonal--uncompute relative-phase Toffoli operations. It agrees in amplitude with the validated circuit on three- and
four-group instances (largest difference $1.5\times10^{-15}$), marks all $4{,}096$ basis states of the full register at the
threshold score $42$ as the validated circuit does (largest deviation $1.0\times10^{-12}$), and its chained routing reproduces
the marginal of the logical circuit on the search register on two- and three-group instances at $j=1$ to $4$ (largest
probability difference $6.1\times10^{-14}$).
\begin{table}[!htb]
  \centering
  \caption{The largest executable iteration count at checkpoint $200$, threshold score $42$, backend \texttt{ibm\_fez}:
  CZ gates per compiled iteration body, the estimate of Eq.~\eqref{eq:jdev-est} and the exact $J_{\mathrm{dev}}$ of
  Eq.~\eqref{eq:jdev}, with the cumulative CZ counts on either side of the limit.}
  \label{tab:jdev}
  \footnotesize
  \begin{tabular}{p{3.5cm}p{2.4cm}p{2.2cm}cp{2.6cm}}
  \toprule
  compilation & status & CZ per body, mean (min--max) & $\widehat J_{\mathrm{dev}}$ & $J_{\mathrm{dev}}$ \\
  \midrule
  validated synthesis, chained layouts & measured, $62$ bodies, CZ counted only & $95{,}246$ ($91{,}445$--$97{,}937$) & $52$ & $52$; $C_{\mathrm{CZ}}(52)=4{,}953{,}241$, $C_{\mathrm{CZ}}(53)=5{,}046{,}837$ \\
  shared-AND synthesis, one iteration compiled whole, simulated backend & prototype, three routing seeds & $61{,}320$--$64{,}050$ & $78$--$81$ & not defined (one iteration only) \\
  shared-AND synthesis, chained layouts, live backend (calibration of 11~November~2026) & measured, $78$ bodies & $64{,}522$ ($61{,}853$--$66{,}987$) & $77$ & $77$; $C_{\mathrm{CZ}}(77)=4{,}966{,}659$, $C_{\mathrm{CZ}}(78)=5{,}032{,}743$ \\
  \bottomrule
  \end{tabular}
\end{table}

\paragraph{Restricted draw.}
With $m_{\mathrm{BBHT}}$ the range of Eq.~\eqref{eq:bbht} at a round, the hardware search draws
\begin{equation}
  j\sim\mathcal U\big\{0,\dots,\min\big(\lceil m_{\mathrm{BBHT}}\rceil-1,\,J_{\mathrm{dev}}\big)\big\}.
  \label{eq:jdraw}
\end{equation}
When $J_{\mathrm{dev}}\ge\lceil m_{\mathrm{BBHT}}\rceil-1$, Eq.~\eqref{eq:jdraw} is the draw of Eq.~\eqref{eq:bbht} and leaves the
random-number stream unchanged; with the rule switched on but inactive, a simulated search reproduces the sequential reference
round for round. When the bound is active, the support of the draw changes, and the search is reported as hardware-constrained
BBHT rather than as an unmodified run of the original algorithm. Criterion D updates its rank distribution after each failed
round with the success probability of Eq.~\eqref{eq:grover} at the iteration count actually executed, so its update is computed
from the executed counts whether or not the bound is active; statements that rely on the unconstrained draw, such as expected
numbers of rounds or of Grover iterations before termination, do not carry over. No out-of-range draw is redrawn, no drawn
circuit is truncated and no simulator outcome replaces a device measurement.

\paragraph{Restricted rounds.}
Over the $R$ executed rounds of a search, the observed fraction of rounds whose draw was restricted is
\begin{equation}
  \widehat p_{\mathrm{cap}}=\frac1R\sum_{\text{rounds}}\mathbf 1\big[J_{\mathrm{dev}}<\lceil m_{\mathrm{BBHT}}\rceil-1\big],
  \label{eq:pcap}
\end{equation}
a count over the executed rounds rather than a probability estimated from hypothetical ones; the search logs the indicator for
every round together with $j$, the upper end of its draw and $J_{\mathrm{dev}}$. The executed values of $j$, the total $Q$ of
Eq.~\eqref{eq:counts}, $\widehat p_{\mathrm{cap}}$ and every active bound with the threshold and the counts that caused it are
reported per checkpoint with the run, in Appendix~\ref{app:ibmloop}.

\paragraph{Service-side checks and evidence.}
The service also limits each qubit to $26.8$ million low-level control-system instructions, to which dynamical decoupling adds
after the gate count, so Eq.~\eqref{eq:jdev} is a necessary screen on the client side and not a guarantee that a circuit is
accepted or executed faithfully; a rejection is logged with the service's error and stops the stage, and the circuit is
neither replaced nor given a smaller $j$. Two jobs of an earlier compilation on \texttt{ibm\_fez} show that the limit binds
this workload: one circuit was rejected for $5{,}010{,}852$ CZ gates against the limit of $5{,}000{,}000$, and a circuit of
$4{,}116{,}057$ CZ gates, run with dynamical decoupling, was accepted and finished; that circuit, with $j=23$, was round $39$ of
the first search. The completed run then executed shared-AND circuits of up to $4{,}284{,}266$ CZ gates ($j=60$), all accepted.

\paragraph{Status and scope.}
Three objects are kept apart: the BBHT schedule of Eq.~\eqref{eq:bbht}, the feasibility bound $J_{\mathrm{dev}}$ of the
compiled circuits, and the trace of the executed counts $j$. The accounting of Eqs.~\eqref{eq:gatecount} and~\eqref{eq:jdev},
the draw of Eq.~\eqref{eq:jdraw} and the logging of Eq.~\eqref{eq:pcap} are implemented; Eq.~\eqref{eq:gatecount} is checked on
assembled circuits, $J_{\mathrm{dev}}$ is measured at one checkpoint, threshold and backend, and the equivalence and routing
checks above are passed.
In the completed run, the restriction was logged on the $191$ rounds from round $40$ of the first search onward, when the search
changed to the shared-AND compilation, and was active in none of them, so every search used the unmodified draw of Eq.~\eqref{eq:bbht}; the largest count executed, $j=60$, stays
below the schedule maximum of $63$ (Section~\ref{sec:res-ibmloop}, Appendix~\ref{app:ibmloop}). The hardware searches use the full Interaction-Guided Schedule
Oracle of Section~\ref{sec:igso}: the integer coefficients from the checkpoint evaluation enter the circuit as constants, the
circuit evaluates the score of every schedule coherently and marks the schedules above the threshold, no table of schedule
scores is computed or loaded, and each measured schedule is checked by one classical evaluation of Eq.~\eqref{eq:igscore}. The
hardware validation of Appendix~\ref{app:provenance}, with a $16$-row lookup oracle, is a separate experiment. These counts
describe feasibility and search behavior on the device and make no statement about end-to-end training acceleration.
\fi

\ifibmloop
\clearpage
\section{Hardware-sampled closed loop: search paths and IBM turnaround}
\label{app:ibmloop}
Each search of Section~\ref{sec:res-ibmloop} ran on the least busy of three $156$-qubit IBM Heron devices at the start of its
checkpoint, with one shot per circuit and up to four circuits per job. Before each search, the compiled oracle was checked on the
constants of that checkpoint, in amplitude on a three-group instance and on all $4{,}096$ basis states of the full register.
Rounds $1$ to $39$ of the first search used earlier compilations of the same oracle; every later round used the shared-AND
compilation of Appendix~\ref{app:ibm-jlimit}. A
round draws its iteration count, runs one circuit and checks the sampled schedule classically with Eq.~\eqref{eq:igscore}; when
a job holds several pre-drawn rounds, its results are read in order and the rounds after an improvement or a stop are not
used. Every sample that entered a decision was measured on the device: no simulator sample replaced a device sample, no
instance was reduced and no draw was changed or truncated.

Table~\ref{tab:ibmloop-paths} gives the threshold path of each search, the rounds and Grover iterations after its last
threshold update, and its circuit with the largest $j$. Table~\ref{tab:ibmloop-cost} gives the IBM turnaround of the jobs, from
submission to completion at IBM with the queue included, split into queue and device time, and the QPU time charged. Over the
five searches, $94$ jobs took $37{,}300$~s ($10.4$~h) of turnaround for $188$~s of charged QPU time. Three further jobs at the
first checkpoint, whose rounds were set aside, took $1{,}768$~s and $6$~s of QPU time, and an acceptance test before the run
took $30$~s and $2$~s.

\begin{table}[!htb]
  \centering
  \caption{Threshold path, stopping phase and circuit with the largest $j$ of each hardware-sampled search. The path lists the
  threshold score $\Phi$ after each update with the round of the update, starting from the uniform-rate baseline at $0$. The
  deepest circuit of the run, round $39$ at checkpoint $200$ ($j=23$, $4{,}116{,}057$ CZ gates, depth $7{,}787{,}203$), used an
  earlier compilation.}
  \label{tab:ibmloop-paths}
  \footnotesize
  \resizebox{\textwidth}{!}{\begin{tabular}{r p{4.3cm} rr rrr}
\toprule
 & & \multicolumn{2}{c}{after the last update} & \multicolumn{3}{c}{circuit with the largest $j$} \\
\cmidrule(lr){3-4}\cmidrule(lr){5-7}
checkpoint & threshold path $\Phi$ (round) & rounds & iterations & $j$ (round) & CZ gates & depth \\
\midrule
$200$ & $0$ $\to3$ (4) $\to5$ (5) $\to9$ (7) $\to11$ (11) $\to19$ (14) $\to42$ (19) & $24$ & $227$ & $54$ ($43$) & $3{,}488{,}695$ & $6{,}088{,}686$ \\
$400$ & $0$ $\to4$ (1) $\to30$ (4) $\to64$ (10) & $25$ & $260$ & $56$ ($34$) & $3{,}896{,}199$ & $6{,}613{,}655$ \\
$600$ & $0$ $\to8$ (3) $\to30$ (4) $\to67$ (23) & $25$ & $209$ & $45$ ($46$) & $3{,}216{,}470$ & $5{,}446{,}316$ \\
$800$ & $0$ $\to6$ (4) $\to12$ (14) $\to18$ (29) & $26$ & $264$ & $60$ ($53$) & $4{,}181{,}458$ & $7{,}119{,}097$ \\
$1000$ & $0$ $\to2$ (14) $\to5$ (22) & $27$ & $247$ & $60$ ($48$) & $4{,}284{,}266$ & $7{,}236{,}444$ \\
\midrule
all & & $127$ of $230$ & $1{,}207$ of $1{,}381$ & & & \\
\bottomrule
\end{tabular}
}
\end{table}

\begin{table}[!htb]
  \centering
  \caption{IBM turnaround of the hardware-sampled searches: jobs with a result, turnaround from submission to completion at
  IBM (queue included), queue and device time, and QPU time charged.}
  \label{tab:ibmloop-cost}
  \footnotesize
  \resizebox{\textwidth}{!}{\begin{tabular}{r l r rr rrr}
\toprule
checkpoint & device & jobs & \multicolumn{2}{c}{IBM turnaround (s)} & queue (s) & device (s) & QPU charged (s) \\
\cmidrule(lr){4-5}
 & & & sum & median / longest job & & & \\
\midrule
$200$ & \texttt{ibm\_fez}, \texttt{ibm\_marrakesh} & $19$ & $6{,}930$ & $261$ / $1{,}213$ & $40$ & $6{,}890$ & $38$ \\
$400$ & \texttt{ibm\_marrakesh} & $17$ & $4{,}755$ & $219$ / $992$ & $28$ & $4{,}727$ & $34$ \\
$600$ & \texttt{ibm\_kingston} & $20$ & $10{,}588$ & $309$ / $3{,}876$ & $4{,}655$ & $5{,}934$ & $40$ \\
$800$ & \texttt{ibm\_kingston} & $20$ & $6{,}406$ & $210$ / $1{,}057$ & $33$ & $6{,}373$ & $40$ \\
$1000$ & \texttt{ibm\_marrakesh} & $18$ & $8{,}620$ & $350$ / $1{,}878$ & $1{,}849$ & $6{,}771$ & $36$ \\
\midrule
all & & $94$ & $37{,}300$ & & $6{,}605$ & $30{,}696$ & $188$ \\
\bottomrule
\end{tabular}
}
\end{table}
\fi

\end{document}